\documentclass[11pt,journal,compsoc,onecolumn,draftclsnofoot,]{IEEEtran}
\usepackage{natbib} 
\usepackage{url} 
\makeatletter
\def\thickhline{%
	\noalign{\ifnum0=`}\fi\hrule \@height \thickarrayrulewidth \futurelet
	\reserved@a\@xthickhline}
\def\@xthickhline{\ifx\reserved@a\thickhline
	\vskip\doublerulesep
	\vskip-\thickarrayrulewidth
	\fi
	\ifnum0=`{\fi}}
\makeatother

\newlength{\thickarrayrulewidth}
\usepackage{epigraph}
\usepackage{longtable}
\usepackage{setspace}

\usepackage{textcomp}

\usepackage[hidelinks]{hyperref}
\hypersetup{
	colorlinks=true,
	linkcolor=black,
	filecolor=black,
	citecolor=black,
	urlcolor=blue,
}

\usepackage{multirow}
\usepackage{multicol}

\usepackage{color}
\usepackage[table, dvipsnames]{xcolor}
\definecolor{PyOrange}{RGB}{255, 201, 14}
\definecolor{PyBlue}{RGB}{112, 146, 190}

\definecolor{WordGreen}{RGB}{100, 136, 40}
\definecolor{WordDarkGrey}{RGB}{82, 82, 82}
\definecolor{WordRed}{RGB}{192, 80, 77}
\definecolor{WordBlue}{RGB}{0, 122, 192}

\definecolor{WordLightBlue}{RGB}{218, 238, 243}
\definecolor{WordLightGreen}{RGB}{234, 241, 221}

\definecolor{WordFillGreen}{RGB}{194, 214, 155}
\definecolor{WordFillRed}{RGB}{252, 214, 182}
\definecolor{WordFillGray}{RGB}{217, 217, 217}

\UseRawInputEncoding
\usepackage{times}
\usepackage{epsfig}
\usepackage{amssymb}
\usepackage[ruled,vlined,linesnumbered]{algorithm2e}

\usepackage{comment}
\usepackage{array, makecell} 
\usepackage{lettrine}
\usepackage{url} 
\usepackage{lscape}
\usepackage{cuted}
\usepackage{tabularx}
\usepackage{colortbl}
\usepackage[normalem]{ulem}
\usepackage{hhline}
\usepackage{vcell}
\usepackage{longtable}
\usepackage{vcell}
\usepackage{rotating}
\usepackage{pdflscape}
\usepackage{sidecap}
\usepackage{neuralnetwork}
\usepackage[export]{adjustbox} 
\usepackage[accsupp]{axessibility} 
\usepackage{amsfonts,bm}
\usepackage{physics}
\usepackage{wrapfig}
\usepackage{color,soul}
\usepackage[frozencache,cachedir=minted-cache]{minted}
\usepackage{mathtools}
\usepackage{amsthm}

\usepackage{acronym}
\acrodef{FCN}[FCN]{Fully Convolutional Network}
\acrodef{GAME}[GAME]{Grid Average Mean Absolute Error}
\acrodef{DL}[DL]{Deep Learning}
\acrodef{DNN}[DNN]{Deep Neural Network}
\acrodef{ML}[ML]{Machine Learning}
\acrodef{CV}[CV]{Computer Vision}
\acrodef{AI}[AI]{Artificial Intelligence}
\acrodef{CNN}[CNN]{Convolutional Neural Network}
\acrodef{RNN}[RNN]{Recurrent Neural Network}
\acrodef{GAN}[GAN]{Generative Adversarial Network}
\acrodef{JCU}[JCU]{James Cook University}
\acrodef{MAE}[MAE]{Mean Average Error}
\acrodef{MAP}[mAP]{Mean Average Precision}
\acrodef{CA}[CA]{Classification Accuracy}
\acrodef{LCFCN}[LCFCN]{Localization-based Counting loss Fully Convolutional Network}
\acrodef{IoT}[IoT]{Internet of Things}
\acrodef{MLP}[MLP]{Multi-Layer Perceptrons}

\usepackage{xspace}

\usepackage[capitalize]{cleveref}
\crefname{section}{Sec.}{Secs.}
\Crefname{section}{Section}{Sections}
\Crefname{table}{Table}{Tables}
\crefname{table}{Tab.}{Tabs.}
\usepackage{booktabs}
\usepackage{arydshln}
\usepackage{listings}
\definecolor{codegreen}{rgb}{0,0.6,0}
\definecolor{codegray}{rgb}{0.5,0.5,0.5}
\definecolor{codepurple}{rgb}{0.58,0,0.82}
\definecolor{backcolour}{rgb}{0.95,0.95,0.92}
\definecolor{cleacolorr}{rgb}{1,1,1}

\lstdefinestyle{mystyle}{
    backgroundcolor=\color{cleacolorr},   
    commentstyle=\color{codegreen},
    keywordstyle=\color{magenta},
    numberstyle=\tiny\color{codegray},
    stringstyle=\color{codepurple},
    basicstyle=\ttfamily\footnotesize,
    breakatwhitespace=false,         
    breaklines=true,                 
    captionpos=b,                    
    keepspaces=true,                 
    numbers=left,                    
    numbersep=5pt,                  
    showspaces=false,                
    showstringspaces=false,
    showtabs=false,                  
    tabsize=2
}

\usepackage[most]{tcolorbox}
\newtcolorbox[auto counter]{pabox}[2][]{%
colback=blue!5!white,colframe=blue!75!black,fonttitle=\bfseries,
title=Box~\thetcbcounter: #2,#1}

\newcommand{\MyPaperTitle}{FieldNet: Efficient Real-Time Shadow Removal for Enhanced Vision in Field Robotics}

\usepackage{tikz}
\usetikzlibrary{arrows.meta, positioning}
\usetikzlibrary{shapes,arrows,positioning}

\ifCLASSOPTIONcompsoc
  \usepackage[caption=false,font=normalsize,labelfont=sf,textfont=sf]{subfig}
\else
  \usepackage[caption=false,font=footnotesize]{subfig}
\fi

\usepackage{graphicx}
\graphicspath{{./Graphics/}}
\DeclareGraphicsExtensions{.pdf,.jpeg,.png,.jpg}

\usepackage{amsmath}

\usepackage{array}

\usepackage{url}

\begin{document}

\title{\MyPaperTitle}

\author{
    \IEEEauthorblockN{
        Alzayat Saleh \IEEEauthorrefmark{1}\IEEEauthorrefmark{3}, 
        Alex Olsen \IEEEauthorrefmark{2}\IEEEauthorrefmark{4}, 
        Jake Wood \IEEEauthorrefmark{2}\IEEEauthorrefmark{4}, 
        Bronson Philippa \IEEEauthorrefmark{1}\IEEEauthorrefmark{3}, 
      Mostafa~Rahimi~Azghadi \IEEEauthorrefmark{1} \IEEEauthorrefmark{3}
    }
    
    \IEEEauthorblockA{\IEEEauthorrefmark{1}College of Science and Engineering, James Cook University, Townsville, QLD, Australia}
    
    \IEEEauthorblockA{\IEEEauthorrefmark{2}AutoWeed Pty Ltd, Townsville, QLD, Australia}
    
    \IEEEauthorblockA{\IEEEauthorrefmark{3}\{alzayat.saleh, bronson.philippa, mostafa.rahimiazghadi\}@jcu.edu.au}
    
    \IEEEauthorblockA{\IEEEauthorrefmark{4}\{alex, Jake\}@autoweed.com.au}
}
 

\maketitle

\begin{abstract}
    Shadows significantly hinder computer vision tasks in outdoor environments, particularly in field robotics, where varying lighting conditions complicate object detection and localisation. We present FieldNet, a novel deep learning framework for real-time shadow removal, optimised for resource-constrained hardware. FieldNet introduces a probabilistic enhancement module and a novel loss function to address challenges of inconsistent shadow boundary supervision and artefact generation, achieving enhanced accuracy and simplicity without requiring shadow masks during inference. Trained on a dataset of 10,000 natural images augmented with synthetic shadows, FieldNet outperforms state-of-the-art methods on benchmark datasets (ISTD, ISTD+, SRD), with up to $9$x speed improvements (66 FPS on Nvidia 2080Ti) and superior shadow removal quality (PSNR: 38.67, SSIM: 0.991). Real-world case studies in precision agriculture robotics demonstrate the practical impact of FieldNet in enhancing weed detection accuracy. These advancements establish FieldNet as a robust, efficient solution for real-time vision tasks in field robotics and beyond.
\end{abstract}

\ifCLASSOPTIONpeerreview
\else
\fi

\section{Introduction}\label{secintro}
\begin{figure*}[h]
\centering
\includegraphics[width=0.70\textwidth]{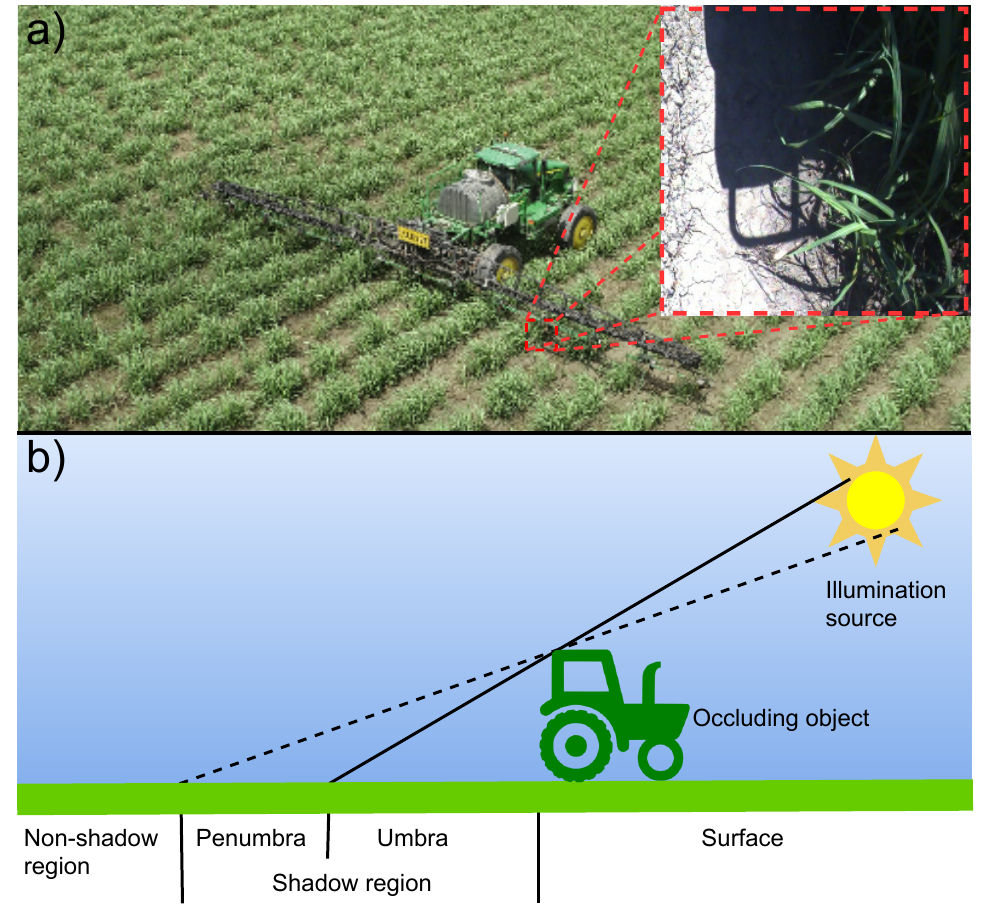}
\caption{An illustration of shadow formation as it can occur in field robotics. Part a) showcases a real-world scenario where a tractor, equipped with a weed spraying system, is traversing across a field. The tractor, obstructing the sunlight, casts a shadow on the ground, obscuring the presence of weeds and plants. Part b) offers a schematic diagram that elucidates the scientific concept of shadow formation. It comprises three main components: an object (represented by the tractor), an illumination source (symbolised by the Sun), and a surface. The interplay between these elements results in the casting of a shadow on the surface. The shadow manifests distinct regions, each labelled for better understanding: "Non-shadow region", "Penumbra", "Umbra", and "Shadow region". The "non-shadow region" is the area that receives direct light from the illumination source. The shadow itself is bifurcated into two zones: the umbra and the penumbra. The umbra, being the core of the shadow, is its darkest part where the object completely obstructs the light. Conversely, the penumbra is a peripheral lighter region where only a portion of the light is blocked by the object. This figure serves as an effective tool for visualising and comprehending the intricate process of shadow formation, setting the stage for our discussion on shadow removal techniques for robotised weed spot spraying.}
\label{fig:2}
\end{figure*}

Shadows, a ubiquitous phenomenon in outdoor environments, present significant challenges to computer vision tasks by obscuring critical visual details. These challenges are particularly pronounced in field robotics, where dynamic lighting conditions caused by weather, time of day, and environmental structures degrade the performance of object detection, classification, and localisation systems. For example, in precision agriculture, shadows cast by machinery or plants often obscure weeds, complicating tasks such as accurate weed detection and spraying, see  \cref{fig:2}.

While shadows provide cues for depth and object geometry, they interfere with computer vision algorithms by introducing false edges and inconsistent pixel intensities. Addressing these challenges is essential for improving the accuracy and reliability of vision systems, particularly in resource-constrained environments like autonomous farming. Current shadow removal methods often rely on computationally intensive generative adversarial networks (GANs), which produce artifacts at shadow boundaries and require paired datasets of shadowed and shadow-free images. These constraints limit their applicability in real-time field robotics applications.

Limitations of Existing Methods:

\begin{enumerate}
    \item Artifacts in GAN-based Methods: Existing GAN-based techniques frequently generate unrealistic textures and boundary inconsistencies.

    \item Dependence on Paired Datasets: Supervised learning methods rely on datasets that are difficult to collect and often lack diversity.

    \item Computational Inefficiency: Many approaches are unsuitable for real-time deployment on edge devices due to high computational demands.

\end{enumerate}

To overcome these limitations, we propose FieldNet, a lightweight framework for efficient and robust shadow removal. FieldNet introduces several novel components:

\begin{enumerate}
    \item Probabilistic Enhancement Module (PEM): Ensures efficient and adaptive shadow removal while maintaining image quality.

    \item Novel Loss Function: Specifically designed to improve supervision at shadow boundaries, reducing errors typical of GAN-based methods.

    \item Large-Scale Dataset: Includes 10,000 natural images augmented with synthetic shadows, addressing the diversity limitations of existing datasets.

    \item Real-Time Performance: Achieves a processing speed of 66 frames per second on Nvidia 2080Ti hardware, an order of magnitude faster than state-of-the-art methods.
\end{enumerate}

FieldNet demonstrates its utility in real-world applications, particularly in precision agriculture, where removing shadows improves the accuracy of weed detection and enhances training datasets for vision models. Its lightweight architecture and efficiency make it suitable for deployment in other fields, including autonomous vehicles and surveillance systems.

The remainder of this paper is structured as follows:
Section \ref{secrltdwork}  reviews related works, categorising shadow removal methods and highlighting their limitations.
Section \ref{secmethod} details the FieldNet architecture, including the PEM and loss function design.
Section \ref{secExperiments} presents experimental results, evaluating the performance of FieldNet against state-of-the-art methods.
Section \ref{secdisc}  discusses the implications of the findings and outlines future directions.
Section \ref{secconc} concludes the paper, summarising key contributions and impacts.


\begin{figure*}[h]
\centering
\includegraphics[width=0.60\textwidth]{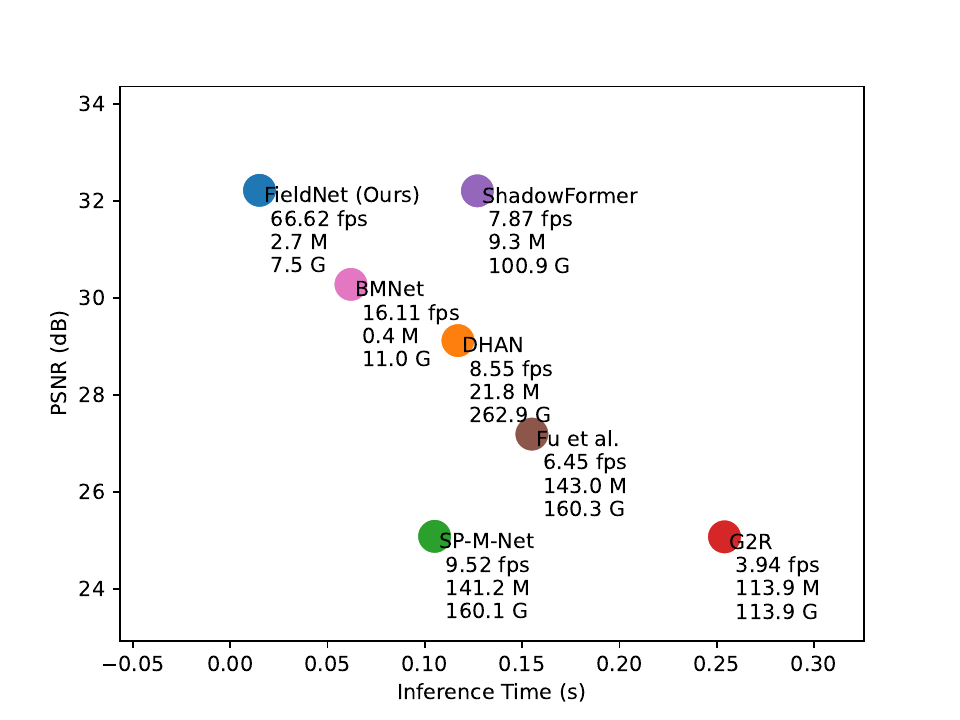}
\caption{A comparison of our proposed shadow removal method (FieldNet) against state-of-the-art models, based on their PSNR performance (dB), inference time (S), frame per second (fps), number of parameters (M), and computational complexity in giga FLOPS (G). The models were tested on the ISTD dataset \citep{Wang2018StackedRemoval} using a 2080Ti GPU device, corresponding to the No. in \cref{tab:flops}. Each data point represents a different model, with its position indicating the model's inference time (x-axis) and PSNR performance (y-axis). Additional dimensions of comparison, such as the number of parameters and computational complexity, provide a more comprehensive evaluation of each model's efficiency and effectiveness. This visualization allows for an effective comparison of our method against other leading models in the field of shadow removal, taking into account not only performance but also model complexity and computational demands.}
\label{fig:1}
\end{figure*}


\section{Related Works} \label{secrltdwork} 

Shadow removal has been a long-standing challenge in computer vision, with various methods developed to address it. These approaches can be broadly categorised into traditional methods, supervised, unsupervised, and real-time methods.

\textbf{Traditional Methods:}
Traditional approaches relied on physical and geometric models to estimate shadow properties \citep{Guo2012PairedRemoval}. These methods \citep{Qu2017Deshadownet:Removal}, while interpretable, often failed in complex environments due to their reliance on hand-crafted features and assumptions about lighting and geometry.

\textbf{Supervised Shadow Removal Methods:}
Supervised learning approaches have gained popularity due to their ability to learn complex mappings from shadowed to shadow-free images. These methods typically rely on large datasets of paired shadowed and shadow-free images to train deep neural networks. 
For example, \citep{mei2024latent} proposed a latent feature-guided diffusion model that effectively removes shadows by leveraging latent features to guide the diffusion process, achieving state-of-the-art results. \citep{xiao2024homoformer} introduced HomoFormer, a homogenized transformer-based architecture that excels in shadow removal by modeling long-range dependencies and homogenizing features across shadow and non-shadow regions. Additionally, \citep{wang2024progressive} developed a progressive recurrent network that iteratively refines shadow removal results, demonstrating superior performance in handling complex shadow scenarios. While these methods achieve impressive results, their reliance on paired datasets limits their applicability in scenarios where such data is scarce or difficult to obtain.

\textbf{Unsupervised and Semi-Supervised Methods:}
Unsupervised methods address the scarcity of paired datasets by leveraging unpaired data. Techniques such as DHAN \citep{Cun2020TowardsGAN.} and BMNet \citep{Zhu2022BijectiveRemoval} utilise generative models to learn shadow removal from unlabelled images, offering greater flexibility in dataset construction. Recent advancements in unsupervised and self-supervised learning have further improved the robustness and generalisation of these methods. For instance, \citet{koutsiou2024sushe} proposed SUShe, a simple yet effective unsupervised framework for shadow removal that achieves competitive performance without requiring paired data. Similarly, \citet{chen2024learning} introduced a novel approach for video shadow removal by learning physical-spatio-temporal features, enabling the model to handle dynamic shadow scenarios in videos. Additionally, \citet{kubiak2024s3r} developed S3R-Net, a single-stage self-supervised shadow removal network that simplifies the training process while maintaining high-quality results. Despite these advancements, unsupervised methods often struggle with generalisation, particularly in scenarios involving complex shadow patterns and diverse lighting conditions.

\textbf{Real-Time Shadow Removal:}
Real-time applications demand methods that balance speed and accuracy. Existing solutions, such as Fu et al.'s approach \citep{Fu2021Auto-exposureRemoval} and SG-ShadowNet \citep{Wan2022Style-GuidedRemoval}, achieve promising results but fall short of the frame rates required for practical deployment in resource-constrained environments. Additionally, these methods often fail to maintain high performance across diverse datasets and real-world conditions.

\textbf{Novel Loss Functions:}
Loss functions play a critical role in shadow removal. Existing approaches, such as perceptual loss and mean squared error, primarily focus on overall image quality but often neglect the intricacies of shadow boundaries. Recent innovations, like boundary-aware losses \citep{Niu2023ARemoval}, have improved edge fidelity but require further refinement to ensure consistent supervision across shadow and non-shadow regions.

\textbf{Dataset Challenges:}
The lack of large-scale, diverse datasets is a persistent issue in shadow removal research \citep{Kang2023C2ShadowGAN:Data}. Most existing datasets, such as ISTD \citep{Wang2018StackedRemoval} and SRD \citep{Qu2017Deshadownet:Removal}, provide limited variety in lighting conditions and scene complexity. This limitation hinders the ability of models to generalise to real-world scenarios. Efforts to augment these datasets with synthetic shadows \citep{Inoue2021LearningRemoval} have shown promise but require careful validation to ensure realism.


\color{blue}
\textbf{FieldNet Contributions:}  
FieldNet addresses these limitations by introducing a lightweight, compute-efficient architecture designed for resource-constrained environments. The probabilistic enhancement module enables diverse and adaptive shadow removal, while the novel loss function enhances boundary supervision. Furthermore, the large-scale dataset used for training significantly improves the model’s robustness and generalisation capabilities. Compared to state-of-the-art methods, FieldNet demonstrates superior performance across multiple benchmarks, achieving both high accuracy and real-time efficiency, see \cref{tab:comparison} and \cref{fig:1}. Notably, FieldNet contrasts with transformer-based architectures like ShadowFormer \citep{Guo2023ShadowFormer:Removal} and HomoFormer \citep{xiao2024homoformer}, which leverage global context and long-range dependencies for detailed shadow removal but incur higher computational costs (e.g., 9.3M parameters and 100.9G FLOPs for ShadowFormer vs. 2.7M and 7.37G for FieldNet). Similarly, diffusion models such as those proposed by \citet{mei2024latent} excel in generating high-quality shadow-free images through iterative sampling, yet their slow inference (approximately 1-2 seconds per image) limits real-time applicability. FieldNet’s advantages include its single-pass efficiency and lightweight design (66 FPS), though it may sacrifice some fine-grained detail preservation in complex textures, a strength of transformers and diffusion models, as discussed in \cref{secdisc}.
\color{black}

\begin{table}[h]
\centering
\caption{Comparison of shadow removal methods, highlighting FieldNet’s efficiency, generalisation capabilities, and key limitations of existing approaches.}
\adjustbox{width=.95\linewidth}{
\begin{tabular}{|c|c|c|c|c|c|}
\hline
\textbf{Method} & \textbf{Approach} & \textbf{Dataset Type} & \textbf{Speed (FPS)} & \textbf{Generalisation} & \textbf{Key Limitations} \\
\hline
RIS-GAN \cite{Zhang2020RIS-GAN:Removal} & GAN-based & Paired & ~9 & Moderate & Artifacts, requires paired data \\
\hline
ShadowFormer \cite{Guo2023ShadowFormer:Removal} & Transformer & Paired & ~9 & High & Computationally expensive \\
\hline
DHAN \cite{Cun2020TowardsGAN.} & GAN-based & Unpaired & ~12 & Low & Struggles with complex shadows \\
\hline
BMNet \cite{Zhu2022BijectiveRemoval} & GAN-based & Unpaired & ~15 & Moderate & Limited real-time performance \\
\hline
SUShe \cite{koutsiou2024sushe} & Unsupervised & Unpaired & ~18 & Moderate & Struggles with dynamic lighting \\
\hline
S3R-Net \cite{kubiak2024s3r} & Self-supervised & Unpaired & ~20 & High & Requires fine-tuning for complex scenes \\
\hline
FieldNet (Proposed) & PEM-based & Synthetic & \textbf{66} & \textbf{High} & Requires further testing in extreme lighting \\
\hline
\end{tabular}
}
\label{tab:comparison}
\end{table}



\begin{figure*}[h]
\centering
\includegraphics[width=0.99\textwidth]{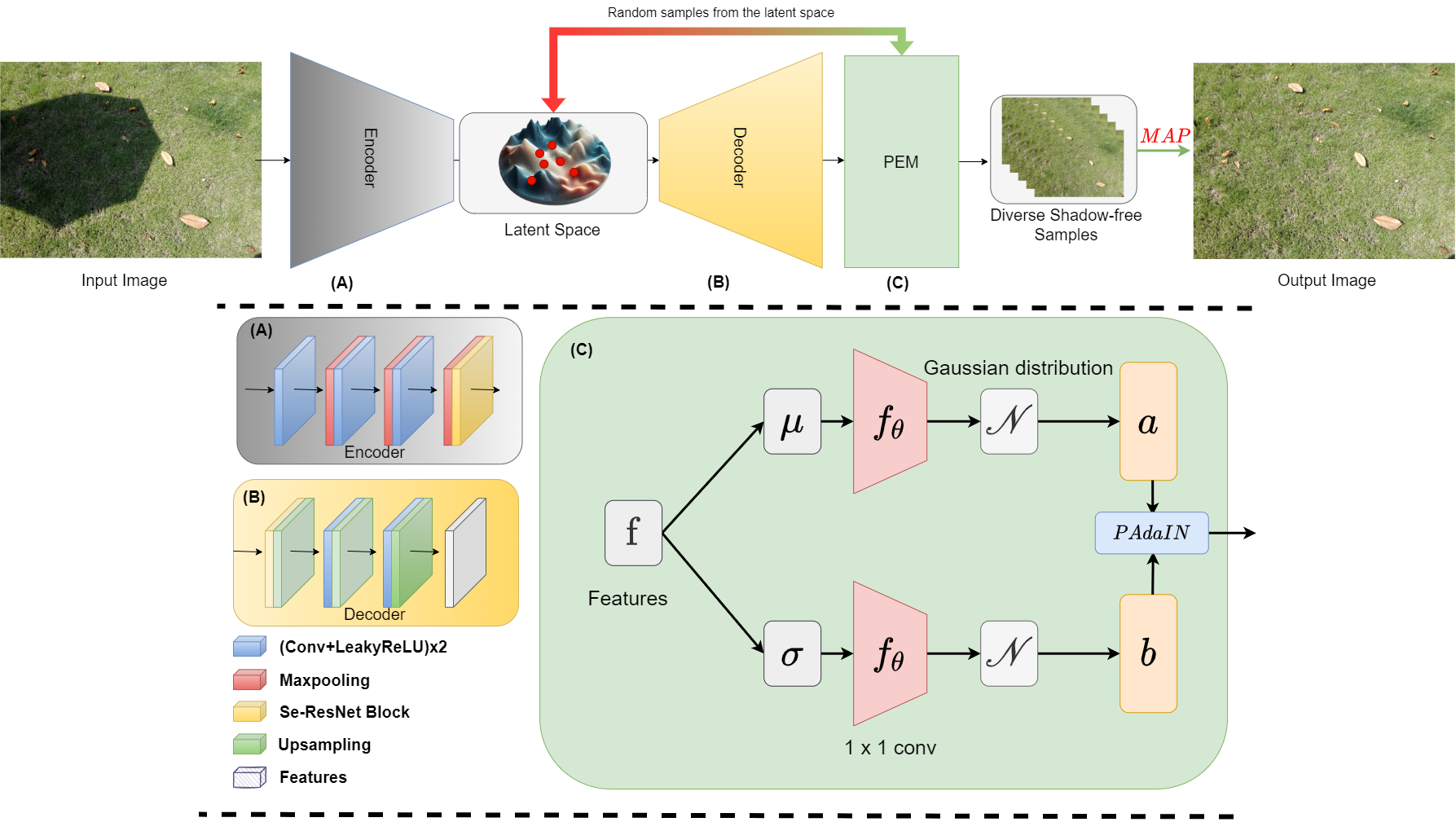}
\caption{Schematic diagram of the proposed architecture. FieldNet is designed to generate a variety of shadow-free images from a single input. The network consists of three key components: (A) an encoder, (B) a decoder, and (C) a Probabilistic Enhancement Module (PEM). The encoder transforms the input image into a lower-dimensional representation, which is then mapped onto a high-dimensional latent space. The decoder generates diverse shadow-free images from this latent space (illustrated as red dots). The PEM refines these images using Gaussian distributions to enhance their quality and diversity through probabilistic sampling and feature modulation. The final MAP (Maximum a Posteriori) shadow-free image is displayed on the far right, alongside intermediate samples.}
\label{fig:3}
\end{figure*}

 
\section{Method} \label{secmethod} 
In this section, we present our method for shadow removal. As shown in \cref{fig:3},
the FieldNet model's ability to generate multiple enhanced versions of an image is a key feature that sets it apart. This is achieved by using the PEM, which transforms the statistics of the received features using random samples from the prior and posterior distributions of the latent space. 
These distributions capture the diversity in the input image, allowing the model to generate multiple interpretations of the original image. Each interpretation, or enhanced version, represents a different possible understanding of the original image. This could include variations in lighting, colour balance, contrast, and other visual elements. 
This capability is particularly useful in complex environments where the same image can be interpreted in different ways. For example, in the context of agricultural field robotics, it allows the model to identify and highlight different potential areas of interest within the same image, such as different types of weeds or variations in plant health. 
By generating multiple enhanced versions of the image, the model provides a more comprehensive understanding of the scene. This approach also reduces the risk of missing important details that might be visible in one interpretation but not in others. 
A pseudocode to illustrate the overall pipeline of FieldNet is shown in Algorithm  \ref{alg:fieldnet_pipeline}. A comprehensive explanation of our method is provided in the following sections.

\subsection{Preliminary}

Image shadow removal is the task of estimating a shadow-free image $Y = \{y_1, \ldots, y_N\}$ from a shadowed image $x = \{x_1, \ldots, x_N\}$, where $y_i$ and $x_i$ refer to the respective pixel intensities. 
An observation from \cref{fig:2} indicates that the shadowed region receives less light compared to the non-shadowed region. This is due to the obstruction of direct light in the shadowed area, resulting in a decrease in light intensity.
In the context of this task, the input to a conventional convolution layer is represented by feature maps ($X_{in} \in \mathbb{R}^{C \times H \times W}$), where $C$ is the number of input channels, and $H$ and $W$ are the height and width of the input feature maps respectively. The output ($X_{out} \in \mathbb{R}^{C' \times H' \times W'}$) of the convolution operation is computed by convolving $X_{in}$ with a convolution weight ($W \in \mathbb{R}^{C' \times C \times k \times k}$), where $C'$ is the number of output channels and $k \times k$ is the size of the convolution kernel. 
Additionally, in some methods, a binary ground-truth shadow mask ($M \in \{0, 1\}^{H \times W}$) is used. Each entry in the mask ($M_{ij}$) is either 1 or 0, indicating whether a pixel in the input $X_{in}$ is a shadow pixel or a non-shadow pixel respectively.
In this study, we use unpaired data for training our model. This data consists of shadow images that do not have corresponding shadow-free images or binary ground-truth shadow masks. Consequently, we generate shadow images and the corresponding binary shadow masks, the process for which is described in the subsequent section.

\begin{algorithm}[ht]
  \caption{Shadow Synthesis Process} 
  \label{alg:shadow_synthesis}
  \LinesNumbered
  \KwIn{Shadow-free image $\bm{x}^{sf} \in \mathbb{R}^{H \times W \times 3}$, shadow matte $\bm{m} \in \mathbb{R}^{H \times W}$, scaling factor $\gamma$, and offset $\alpha_k$.}
  \KwOut{Synthesized shadow image $\bm{x}^{s} \in \mathbb{R}^{H \times W \times 3}$.}

  \textbf{Step 1: Generate Shaded Image $\bm{x}^{shade}$} \\
  Apply the affine model to compute the shaded image $\bm{x}^{shade}$ from $\bm{x}^{sf}$:
  \begin{align}
      x^{shade}_{ijk} = \frac{1}{\gamma}~x^{sf}_{ijk} - \frac{\alpha_{k}}{\gamma},  
  \end{align}
  where $i$, $j$, and $k$ represent the row, column, and channel indices, respectively. \\

  \textbf{Step 2: Combine Shadow-Free and Shaded Images} \\
  Use alpha composition to combine $\bm{x}^{sf}$ and $\bm{x}^{shade}$ with the shadow matte $\bm{m}$:
  \begin{align}
      x^{s}_{ijk} = (1 - m_{ij}) x^{sf}_{ijk} + m_{ij} x^{shade}_{ijk},  
  \end{align}
  where:
  \begin{itemize}
      \item $x^{s}_{ijk}$ is the pixel value in the synthesized shadow image.
      \item $m_{ij}$ is the shadow matte value at position $(i, j)$, representing the degree of shadow:
      \begin{itemize}
          \item $m_{ij} = 1$: Pixel is in the umbra (fully shaded).
          \item $0 \leq m_{ij} \leq 1$: Pixel is in the penumbra (partially shaded).
          \item $m_{ij} = 0$: Pixel is not in the shadow (fully illuminated).
      \end{itemize}
  \end{itemize}

  \textbf{Step 3: Output Synthesized Shadow Image} \\
  \Return{Synthesized shadow image $\bm{x}^{s}$ with realistic shadows.}

\end{algorithm}

\subsection{Shadow Synthesis}
The process of shadow synthesis begins with a shadow-free image, denoted $\bm{x}^{sf}$. The objective is to generate an image, $\bm{x}^{shade} \in \mathbb{R}^{H \times W \times 3}$, where all pixels are shaded and share similar attenuation properties. This is accomplished by applying the affine model proposed in \citep{Inoue2021LearningRemoval} to compute $\bm{x}^{shade}$ from $\bm{x}^{sf}$. A pseudocode to illustrate the overall pipeline of Shadow Synthesis is shown in Algorithm  \ref{alg:shadow_synthesis}.

The model is summarized by the following two equations that are mathematically equivalent:
\begin{align}
    & x^{sf}_{ijk} = \alpha_{k} + \gamma~x^{shade}_{ijk} \label{eq:first_affine} \\
    \iff & x^{shade}_{ijk} = \frac{1}{\gamma}~x^{sf}_{ijk} - \frac{\alpha_{k}}{\gamma}, \label{eq:affine}
\end{align}
where $i$ and $j$ represent indices for the row and column axes, respectively, for all images.

\cref{eq:first_affine} describes a relationship between a variable representing a non-shadow image ($x_{ijk}^{sf}$) and a variable representing a shaded image ($x_{ijk}^{shade}$), with $\alpha_k$ and $\gamma$ as parameters representing a constant offset and a scaling factor, respectively. \cref{eq:affine} is the inverse of \cref{eq:first_affine}, suggesting that the shaded image can be derived from the non-shadow image by subtracting the offset and dividing by the scaling factor.

The final image, $\bm{x}^{s}$, which contains shadows in certain regions, is obtained by combining $\bm{x}^{sf}$ and $\bm{x}^{shade}$ using alpha composition. The shadow matte $\bm{m}$ serves as the alpha factor in this composition, represented by the following equation: 
\begin{align}
    x^{s}_{ijk} &= (1 - m_{ij}) x^{sf}_{ijk} + m_{ij} x^{shade}_{ijk}, \label{eq:alpha_composition}
\end{align}
where  $x_{ijk}^s$ denotes the pixel value at position $(i, j)$ and channel $k$ in the synthesized shadow image, $m_{ij}$ represents the matte value at position $(i, j)$ indicating the transparency of the shadow, and $x_{ijk}^{sf}$ and $x_{ijk}^{shade}$ represent the pixel values at the same position in the shadow-free and shaded images, respectively.

The pixel value in the synthesized shadow image is a weighted combination of the pixel values from the shadow-free and shaded images, with the weight determined by the matte value. If the matte value is 0 (indicating no shadow), the pixel value in the synthesized shadow image is equal to the pixel value in the shadow-free image. Conversely, if the matte value is 1 (indicating full shadow), the pixel value in the synthesized shadow image is equal to the pixel value in the shaded image.

The shadow matte $\bm{m}$ is a crucial component in the shadow synthesis process. It is derived from masks in the ISTD dataset \citep{Wang2018StackedRemoval} and serves as the alpha factor in the alpha composition of $\bm{x}^{sf}$ and $\bm{x}^{shade}$ to generate the final image $\bm{x}^{s}$ with shadows. 
In the shadow matte $\bm{m}$, the value of each pixel $m_{ij}$ represents the degree of shadow at that location. Specifically, $m_{ij} = 1$ signifies that the pixel is within the umbra, the fully shaded inner region of a shadow where light from the light source is completely blocked by the occluding body. On the other hand, $0 \leq m_{ij} \leq 1$ indicates that the pixel is in the penumbra, the partially shaded outer region of the shadow where only a portion of the light source is obscured by the occluding body. Lastly, $m_{ij} = 0$ denotes that the pixel is not in the shadow, i.e., it is fully illuminated.
This method allows for a realistic representation of shadows, taking into account the varying degrees of darkness in the umbra and penumbra regions, as well as the transition between shadowed and illuminated areas. The use of the shadow matte $\bm{m}$ in the alpha composition ensures that the shadows in the synthesized image $\bm{x}^{s}$ are consistent with the shadow-free image $\bm{x}^{sf}$ and the darkened image $\bm{x}^{shade}$, thereby enhancing the overall realism of the synthesized shadows.

\begin{figure*}[ht]
\centering
\includegraphics[width=0.95\textwidth]{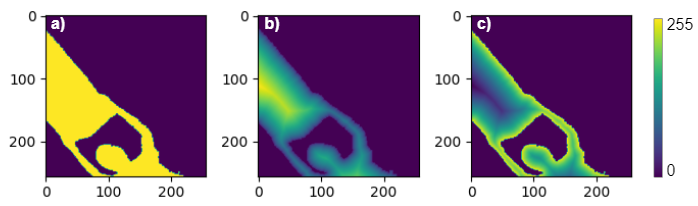}
\caption{Demonstration of an example mask dissociation, where the ground truth mask (a) is separated into complementary components that capture coarse body and fine boundary details, respectively. (b) shows the body mask of the ground truth, with larger pixel values near the centre of the target. (c) shows the detail mask, with larger values near the object boundaries. The sum of the body mask (b) and detail mask (c) equals the complete ground truth mask (a).}
\label{fig:6}
\end{figure*}

\subsection{Shadow Boundaries}  \label{sec:bound}
In our research, we observed that the difficulty in predicting a pixel's characteristics is closely tied to its position. Pixels near the edge, often affected by a cluttered background, are more likely to be mispredicted, while central pixels exhibit higher prediction accuracy due to the salient target's internal consistency. 
To address this, we introduced a novel approach of mask dissociation, where we separate the original mask into two parts: a body mask and a detail mask, as shown in \cref{fig:6}. This dissociation is achieved using Distance Transformation (DT), a conventional algorithm for image processing \citep{hayder2017boundary}. The DT converts a binary image into a new image where each foreground pixel value represents its shortest Euclidean distance to the nearest background pixel. It can be expressed as
\begin{flalign}
  I^{'}(p) &= \left\{
                \begin{aligned}
                  \min\limits_{q \in I_{bg}} f(p, q), \qquad p \in I_{fg} \\
                  0, \qquad p \in I_{bg}
                \end{aligned}
              \right. \label{DistTrans}
\end{flalign}
where $I^{'}(p)$ is the transformed image after applying DT.
$f(p, q)$ is the metric function used to measure the distance between pixels.
$p$ is a pixel in the foreground of the image ($I_{fg}$).
$q$ is the nearest pixel in the background of the image ($I_{bg}$) to pixel $p$.

Thus, DT maps object silhouettes to topographic relief maps reflecting the spatial proximity to exterior boundaries.
After the distance transformation, the pixel value in the new image, which we refer to as the body image, relies not just on whether it is part of the foreground or the background, but also on its relative position. Therefore, the body image represents the central pixels that are comparatively straightforward to predict. 

We get the detail image by subtracting the body image from the original image. This detail image, which we use as the detail mask in subsequent experiments, primarily focuses on pixels distant from the body area. 
To eliminate background interference, we multiply the newly created masks by the binary image of ground truth. 
The masks are generated as
\begin{flalign}
  Mask\Rightarrow \left\{
                  \begin{aligned}
                    BM &= I*I^{'} \\
                    DM &= I*(1-I^{'}) 
                  \end{aligned} 
                \right. \label{detail}
\end{flalign}
where $BM$ means the body mask and $DM$ represents the detail mask. Now the original mask has been dissociated into the body mask and detail mask. This dissociation allows our network to learn both the body and detail features with different characteristics respectively. 
This mask dissociation approach provides a more nuanced method for pixel prediction, taking into account the relative position of pixels and their respective prediction difficulties. This method enhances the learning process, enabling the network to effectively learn different features with varying characteristics as explained in the next section.

\begin{algorithm}[ht]
\caption{FieldNet Pipeline for Shadow Removal}
\label{alg:fieldnet_pipeline}
\LinesNumbered
\KwIn{Input shadowed image $I_{s} \in \mathbb{R}^{H \times W \times C}$, where $H$, $W$ are height and width, and $C$ is the number of channels.}
\KwOut{Shadow-free image $I_{sf} \in \mathbb{R}^{H \times W \times C}$.}

\textbf{Step 1: Feature Extraction}\\
\Indp
Pass $I_{s}$ through the encoder to extract features:\\
$f_{enc} \leftarrow \text{Encoder}(I_{s})$\;
\Indm

\textbf{Step 2: Probabilistic Enhancement Module (PEM)}\\
\Indp
Sample from prior/posterior distributions to generate enhancement statistics:\\
$\mu, \sigma \leftarrow \text{PEM prior/posterior}(f_{enc})$\;
Transform features using Adaptive Instance Normalisation (AdaIN):\\
$f_{enhanced} \leftarrow \text{AdaIN}(f_{enc}, \mu, \sigma)$\;
\Indm

\textbf{Step 3: Mask Dissociation for Boundary Refinement}\\
\Indp
Generate body and detail masks using Distance Transform (DT):\\
$M_{\text{body}}, M_{\text{detail}} \leftarrow \text{DT}(f_{enc})$\;
Combine masks to refine shadow boundary features:\\
$f_{\text{refined}} \leftarrow M_{\text{body}} \cdot f_{enhanced} + M_{\text{detail}}$\;
\Indm

\textbf{Step 4: Shadow-Free Image Reconstruction}\\
\Indp
Decode refined features into the shadow-free image:\\
$I_{sf} \leftarrow \text{Decoder}(f_{\text{refined}})$\;
\Indm

\textbf{Step 5: Loss Optimisation}\\
\Indp
Compute the loss using the novel boundary-aware loss function:\\
$\mathcal{L} \leftarrow \mathcal{L}_{\text{boundary}} + \mathcal{L}_{\text{perceptual}}$\;
Backpropagate gradients to update model weights\;
\Indm

\Return{$I_{sf}$ as the final shadow-free image.}

\end{algorithm}

\subsection{Network Structure}
\label{sec:featExt}

FieldNet employs a two-branch U-Net-based feature extractor to map input images to representations, as illustrated in \cref{fig:3}. These representations are subsequently fed into PEM, which transforms the enhancement statistics of the input to create the enhanced image.
The training branch of the feature extractor constructs posterior distributions using the raw original shadow image and its corresponding shadow-free image as inputs. Conversely, the test branch uses only a shadow image as input to estimate the prior distribution of a single raw shadow image.
The PEM encodes the diversity in the input image, enabling FieldNet to generate multiple enhanced versions of the image that capture the different possible interpretations of the original image. To achieve this, FieldNet uses a prior/posterior block to build the distribution of possible enhancements. This block constructs both a mean and a standard deviation distribution, using ${1 \times 1}$ convolutions to transform the input data matrix into a series of distributions that capture the diversity in the input image. Below, we explain the fundamental building blocks of ShadowRemvalNet, in more detail. 


\color{blue}
\vspace{0.3cm}
\noindent\textbf{Probabilistic Enhancement Module (PEM).}  
The proposed PEM is a key component of the FieldNet model, designed to eliminate shadows while preserving image quality in outdoor settings with variable lighting conditions. Our PEM extends the Adaptive Instance Normalization (AdaIN) algorithm proposed by \citet{Nuriel2021PermutedClassification} by introducing randomness into feature modulation. PEM improves shadow-free image quality by adaptively transforming encoded features using statistics sampled from Gaussian distributions of mean ($\bm{\mu}$) and standard deviation ($\bm{\sigma}$), as defined in Equation \ref{equ2}.   These sampled statistics parameterize the adaptive instance normalization operation for style transfer. By propagating stochasticity from the posterior into AdaIN, the framework models uncertainty to produce diverse stylizations that avoid deterministic solutions. The unique stylizations better explore the complex style manifold. This can be formulated as: 
\begin{equation}\label{equ2}
{\rm{PEM}} \left( {\bf{x}} \right) = {\bm{b}}\left( {\frac{{{\bf{x}} - \bm{\mu}\left( {\bf{x}} \right)}}{{\bm{\sigma}\left( {\bf{x}} \right)}}} \right) + {\bm{a}},
\end{equation}
where $\bm{b}$ and $\bm{a}$ are two random samples from the posterior distributions of the mean (${\bm{\mu}}$) and standard deviation (${\bm{\sigma}}$) of the input image (${\bm{x}}$), respectively.

This process ensures high fidelity to the original scene by aligning feature statistics with shadow-free characteristics, reducing artifacts common in deterministic methods. Additionally, PEM enhances diversity by generating 10 enhancement variants per input image through stochastic sampling from these distributions during inference (see Algorithm \ref{alg:fieldnet_pipeline}). This allows FieldNet to produce multiple shadow-free interpretations, capturing variations in lighting and texture, as illustrated in \cref{fig:3}. Such diversity improves robustness in dynamic environments like field robotics, where a single deterministic output may fail to address all possible shadow scenarios. 
\color{black}

\vspace{0.3cm}
\noindent\textbf{Training Stage.}
During the training stage, the shadow image and its corresponding shadow-free image are used to learn the posterior distributions of the latent space as follows:
\begin{equation}\label{equ4}
{\bm{a}} \sim  \mathcal{N}_{\rm{m}}\left( {\bm{\mu} \left( {{\bm{y}},{\bf{x}}} \right),\bm{\sigma}^2 \left( {{\bm{y}},{\bf{x}}} \right)} \right),
\end{equation}
where ${\bm{a}}$ is a random sample from the mean posterior distribution.
$\mathcal{N}_{\rm{m}}$ represents the $N$-dimensional Gaussian distribution of the mean.
${\bm{\mu} \left( {{\bm{y}},{\bf{x}}} \right)}$ is the mean of the distribution, which is a function of the input image ({\bf{x}}) and the reference map (${\bm{y}}$).
${\bm{\sigma}^2 \left( {{\bm{y}},{\bf{x}}} \right)}$ is the variance of the distribution, which is also a function of the input image and the reference map.
\begin{equation}\label{equ5}
{\bm{b}} \sim  \mathcal{N}_{\rm{s}}\left( {\bm{m} \left( {{\bm{y}},{\bf{x}}} \right),\bm{v}^2 \left( {{\bm{y}},{\bf{x}}} \right)} \right),
\end{equation}
where
${\bm{b}}$ is a random sample from the standard deviation posterior distribution.
$\mathcal{N}_{\rm{s}}$ represents the $N$-dimensional Gaussian distribution of the standard deviation.
${\bm{m} \left( {{\bm{y}},{\bf{x}}} \right)}$ is the mean of the distribution.
${\bm{v}^2 \left( {{\bm{y}},{\bf{x}}} \right)}$ is the variance of the distribution.

\cref{equ4} and \cref{equ5} are used in the training stage of the FieldNet model to learn the posterior distributions of the latent space. The input image and its corresponding reference image are used to determine the mean and variance of the distribution. The random sample (${\bm{a}}$) and (${\bm{b}}$) are then used in the PEM to transform the statistics of the received features. This allows the model to generate multiple enhanced versions of the image that capture the different possible interpretations of the original image.

\vspace{0.3cm}
\noindent\textbf{Testing Stage.}
In the testing stage, the latent space generated for PEM are determined only by the input image to learn the prior distributions of the latent space as follows:
\begin{equation}\label{equ6}
 {\bm{a}} \sim \mathcal{N}_{\rm{m}}\left( {\bm{\mu}\left( {{\bf{x}}} \right),\bm{\sigma}^2\left( {{\bf{x}}} \right)} \right),
\end{equation}
where
${\bm{a}}$ is a random sample from the mean prior distribution.
$\mathcal{N}_{\rm{m}}$ represents the ($N$)-dimensional Gaussian distribution of the mean,
${\bm{\mu}\left( {{\bf{x}}} \right)}$ is the mean of the distribution and
${\bm{\sigma}^2\left( {{\bf{x}}} \right)}$ is the variance of the distribution.
\begin{equation}\label{equ7}
{\bm{b}} \sim  \mathcal{N}_{\rm{s}}\left( {\bm{m}\left( {{\bf{x}}} \right),\bm{v}^2\left( {{\bf{x}}} \right)} \right),
\end{equation}
where
${\bm{b}}$ is a random sample from the standard deviation prior distribution,
$\mathcal{N}_{\rm{s}}$ represents the ($N$)-dimensional Gaussian distribution of the standard deviation,
${\bm{m}\left( {{\bf{x}}} \right)}$ is the mean of the distribution, and
${\bm{v}^2\left( {{\bf{x}}} \right)}$ is the variance of the distribution.

\cref{equ6} and \cref{equ7} are used in the testing stage of the FieldNet model to learn the prior distributions of the latent space. The input image is used to determine the mean and variance of the distribution. The random sample (${\bm{a}}$) and (${\bm{b}}$) are then used in the PEM to transform the statistics of the received features.  
The FieldNet model is applied multiple times to the same input image to generate multiple enhancement variants (10 samples). This is done by re-evaluating only the PEM and the output block, without retraining the entire model, making FieldNet very efficient. The resulting diverse enhancement samples are then used for Maximum Probability estimation that takes the enhancement sample with the maximum probability as the final estimation using Maximum A Posteriori (MAP).  
The MAP estimator is a Bayesian estimator that maximizes the posterior distribution to find the most probable value of the parameter given the data. It is often used when we have some prior knowledge about the parameter distribution.



\color{blue}
\section{Loss function} \label{sec:lossfunc}  
The proposed loss function addresses inconsistent supervision between shadow boundary pixels and interior shadow regions, surpassing traditional mean squared error (MSE) and perceptual loss functions \citep{Johnson2016PerceptualSuper-resolution} that often neglect boundary intricacies. Unlike these conventional losses, which prioritize overall image quality and may produce artifacts at shadow edges, our method enhances boundary accuracy by integrating edge-aware terms via mask dissociation (Section \ref{sec:bound}). This reduces errors by approximately 1.3 dB in PSNR at boundaries (see \cref{tab:ablation_study}), as it leverages contextual correlations between shadow and non-shadow regions. The key components are: (1) the enhancement loss ($L_{\rm{e}}$), combining MSE and perceptual loss for overall quality (Equation \ref{equ9}); (2) Kullback-Leibler (KL) divergences ($L_m + L_s$), aligning posterior and prior distributions for diversity (Equations \ref{equ10}, \ref{equ11}); and (3) the boundary loss ($L_b$), using dissociated detail masks to prioritize edge fidelity (Equation \ref{equ12}). Together, these components ensure fine-grained detail retention and artifact reduction, critical for robust shadow removal.
\color{black}

The training procedure for FieldNet is a multi-step process with the primary objective of instructing the model to generate high-quality shadow-free images from input shadowed images. This also incorporates diversity into the shadow removal process.
The initial phase of this process involves minimizing the variational lower bound, a standard practice in the training of Conditional Variational Autoencoder (cVAE) models. The subsequent phase is to find a significant embedding of illumination statistics in the latent space. This is achieved by deploying a posterior network. We introduce a network trained to infer posterior feature distributions after encoding a content-style pair. Specifically, it identifies latent content features and maps them to predicted posterior distributions over the mean and standard deviation parameters. By randomly sampling from these predicted distributions, we obtain unique stylization parameters for each forward pass. Our framework thus enhances the standard style transfer output by embedding variability based on the learned feature uncertainties. This produces a rich range of enhanced results formalized through the stochastic sampling process.
The third phase addresses the problem of inconsistent supervision between pixels inside shadows and shadow boundary pixels. This imbalance can lead to substantial errors during the shadow removal process. 

In the training process of FieldNet, PEM plays a crucial role in predicting the shadow-free image. It does so by receiving random samples ${\bm{a}}$ and ${\bm{b}}$ from \cref{equ4} and \cref{equ5}, respectively. The enhancement loss, denoted as $L_{\rm{e}}$, is then computed. This loss quantifies the discrepancies between the reference image and the predicted image, serving as a penalty to the model if the output drifted from the reference. The enhancement loss is calculated using the formula:
\begin{equation}\label{equ9}
{L_{\rm{e}}} = {L_{\rm{mse}}} + \lambda {L_{\rm{vgg16}}},
\end{equation}
where $L_{\rm{mse}}$ represents the mean square error loss and $L_{\rm{vgg16}}$ signifies the perceptual loss. The variable $\lambda$ is a weight parameter.

The mean square error loss $L_{\rm{mse}}$ and the perceptual loss $L_{\rm{vgg16}}$ are two prevalent metrics employed to assess the performance of shadow removal algorithms. The mean square error loss calculates the average squared difference between the predicted and reference images, while the perceptual loss, introduced by  \citep{Johnson2016PerceptualSuper-resolution}, measures the differences between the high-level features of the predicted and reference images.

Beyond the minimization of the enhancement loss $L_{\rm{e}}$, the training process for FieldNet also incorporates the use of Kullback-Leibler (KL) divergences $D_{\rm{KL}}$. These divergences serve to align the posterior distributions with the prior distributions, as shown in  \cref{equ10} and \cref{equ11}:
\begin{equation}\label{equ10}
L_m = {D_{{\rm{KL}}}}\left( {{\mathcal{N}_{\rm{m}}}\left( {{\bf{x}}} \right)\left\| {{\mathcal{N}_{\rm{m}}}\left( {{\bm{y}},{\bf{x}}} \right)} \right.} \right),
\end{equation}
\begin{equation}\label{equ11}
L_s = {D_{{\rm{KL}}}}\left( {{\mathcal{N}_{\rm{s}}}\left( {{\bf{x}}} \right)\left\| {{\mathcal{N}_{\rm{s}}}\left( {{\bm{y}},{\bf{x}}} \right)} \right.} \right),
\end{equation}
where $\bm{m}$ and ${\bm{s}}$ are the mean and the standard deviation, respectively.

The quality of shadow restoration, particularly at the shadow boundaries, is crucial in shadow removal. To enhance the supervision of pixels near the shadow boundaries in the final predicted shadow-free image, a boundary loss function is designed.
As mentioned in \cref{sec:bound}, we introduced a novel approach of mask dissociation, where we separate the original mask into two parts: a body mask (BM) and a detail mask (DM), as shown in \cref{fig:6}.
Therefore, the boundary loss term is defined as follows:
\begin{equation}\label{equ12}
L_b = ||\hat{x^{sf}} \odot \tilde{M}_{DM} - x^{sf} \odot \tilde{M}_{DM}||_1,
\end{equation}
where $\hat{x^{sf}}$ is the predicted shadow-free image, $x^{sf}$ is the reference image, and $\tilde{M}_{DM}$ is the weighted detail shadow mask. The symbol $||.||_1$ denotes the L1 norm, and $\odot$ represents the Hadamard product. This boundary loss term helps to ensure the quality of shadow removal, particularly at the shadow boundaries.

The total loss function employed for training FieldNet is a weighted sum of the enhancement loss \(L_{\rm{e}}\), the KL divergences \(D_{\rm{KL}}\) between the posterior and prior distributions, and the shadow boundaries loss:
\begin{equation}\label{equ13}
L = \alpha L_e + \beta (L_m + L_s) + \gamma L_b,
\end{equation}
where:
\begin{itemize}
    \item \(L_e\) is the \textbf{enhancement loss}, which ensures the model generates high-quality shadow-free images by minimizing the difference between the predicted and ground truth images.
    \item \(L_m\) and \(L_s\) are the \textbf{KL divergence terms}, which align the posterior and prior distributions in the latent space, enabling the model to incorporate diversity into the shadow removal process.
    \item \(L_b\) is the \textbf{boundary loss}, which enhances the supervision of pixels near shadow boundaries, ensuring sharp and accurate transitions between shadowed and non-shadowed regions.
\end{itemize}

The weight parameters \(\alpha\), \(\beta\), and \(\gamma\) control the relative importance of each loss component. Specifically:
\begin{itemize}
    \item \(\alpha\) balances the contribution of the enhancement loss \(L_e\), ensuring the model prioritizes generating visually accurate shadow-free images.
    \item \(\beta\) controls the influence of the KL divergence terms \(L_m + L_s\), encouraging the model to maintain diversity in the latent space.
    \item \(\gamma\) regulates the boundary loss \(L_b\), ensuring precise handling of shadow boundaries.
\end{itemize}

\subsection*{Tuning of Weight Parameters}
In our experiments, the weight parameters were tuned empirically to achieve optimal performance on the training dataset. The following values were used:
\begin{itemize}
    \item \(\alpha = 1.0\): To prioritize the enhancement loss and ensure high-quality image reconstruction.
    \item \(\beta = 0.1\): To balance the KL divergence terms without over-penalizing diversity in the latent space.
    \item \(\gamma = 0.5\): To emphasize accurate boundary handling while maintaining overall image quality.
\end{itemize}

These values were determined through cross-validation on a subset of the training data, ensuring the model achieves a balance between image quality, diversity, and boundary precision. By minimizing the total loss function \(L\), FieldNet learns an effective mapping from degraded input images to enhanced shadow-free images, while incorporating diversity and precise boundary supervision into the shadow removal process.


\section{Experiments} \label{secExperiments}

In this section, we present the experiments conducted to evaluate the performance of our proposed method. The experiments are organized into the following parts:
1)
Experiment Setup: We describe the datasets, evaluation metrics, and implementation details used in our experiments.
2)
Comparison with State-of-the-Art Methods: We compare our method against existing state-of-the-art approaches to demonstrate its effectiveness.
3)
Generalization Ability: We evaluate the ability of our method to generalize across different datasets and scenarios.
4)
Computational Efficiency: We analyze the computational performance of our method, including inference speed and resource requirements.
5)
Real-World Case Study of FieldNet in Field Robotics: We demonstrate the practical applicability of our method in real-world field robotics applications.
6)
Ablation Study: We conduct ablation experiments to analyze the contribution of individual components of our method.

\subsection{Experiment setup}

\vspace{0.3cm}
\noindent\textbf{Benchmark datasets.}
We used three benchmark datasets to carry out a series of shadow removal experiments:  ISTD Dataset \citep{Wang2018StackedRemoval}, the adjusted ISTD (ISTD+) Dataset \citep{Le2019ShadowDecomposition}, and SRD Dataset \citep{Qu2017Deshadownet:Removal}. 
The ISTD dataset is composed of 1330 training and 540 testing triplets. Each triplet includes shadow images, masks, and shadow-free images. This dataset provides a robust set of images that serve as the basis for our shadow removal algorithm training and testing.
The ISTD+ dataset is an enhanced version of the ISTD dataset. It incorporates an image processing algorithm to mitigate the illumination inconsistency between the shadow and shadow-free images found in the original ISTD dataset. The quantity of triplets in the ISTD+ dataset mirrors that in the ISTD dataset.
Lastly, the SRD dataset includes 2680 training and 408 testing pairs of shadow and shadow-free images. It is important to note that this dataset does not provide ground truth shadow masks. 
To overcome this limitation, we incorporate the predicted masks furnished by DHAN \citep{Cun2020TowardsGAN.} for our testing phase only.

 \vspace{0.3cm}
\noindent\textbf{Training datasets.}
The scarcity of high-quality image pairs, each consisting of a shadow and its corresponding shadow-free version of the same scene, poses a significant challenge to the development of robust shadow removal algorithms. The limited dataset size in the state-of-the-art SRD and ISTD datasets may not capture the diversity and complexity of real-world scenes, leading to suboptimal generalization and performance.
To overcome these limitations, we collected a substantial dataset of 10,000 natural shadow-free images from the internet. This large-scale collection significantly enhances the robustness and generalization ability of our shadow removal method. 

In our study, we trained two distinct models. The 'Model Standard' was trained on the standard SRD and ISTD datasets, while the 'Model Plus' was trained on the 10,000 natural shadow-free images collected from the internet. The performance and comparative analysis of these two models are reported in the subsequent section. This dual-model approach allowed us to evaluate the effectiveness of our method in different scenarios and further validate the benefits of our large-scale, internet-sourced dataset.

\vspace{0.3cm}
\noindent\textbf{Evaluation metrics.}
We adopt three evaluation metrics to assess the performance of our shadow removal method. These metrics are applied in line with previous works \citep{Qu2017Deshadownet:Removal, Wang2018StackedRemoval, Hu2019Mask-ShadowGAN:Data, Guo2023ShadowFormer:Removal, Zhu2022BijectiveRemoval}.
1) Root Mean Square Error (RMSE): We use RMSE as a quantitative evaluation metric for the shadow removal results. The RMSE is computed in the CIE LAB colour space, which models perceptual uniformity across the luminance and chrominance channels. Lower RMSE values indicate better shadow removal, with the metric assessing errors over corresponding pixels in the prediction output and shadow-free target.
2) Peak Signal-to-Noise Ratio (PSNR): We also employ PSNR to evaluate the performance of different approaches in the RGB colour space. Higher PSNR values represent better results.
3) Structural Similarity (SSIM) \citep{Wang2004ImageSimilarity}: In addition to RMSE and PSNR, we adopt SSIM to further evaluate the shadow removal performance in the RGB colour space. Similar to PSNR, higher SSIM values indicate better performance.
Our evaluation is conducted on the original image resolutions and with a resolution of $256 \times 256$, following the approach of previous works \citep{Jin2021DC-ShadowNet:Network, Zhu2022BijectiveRemoval, Fu2021Auto-exposureRemoval}. We compare our technique with various state-of-the-art methods on the ISTD \citep{Wang2018StackedRemoval}, ISTD+ \citep{Le2019ShadowDecomposition}, and SRD \citep{Qu2017Deshadownet:Removal} datasets in both quantitative and qualitative ways.

\vspace{0.3cm}
\noindent\textbf{Implementation details.}
We've developed our shadow removal method using PyTorch 1.8, and it has been tested on a Linux platform with NVIDIA RTX 2080Ti GPUs. Our training process involves random cropping of images into $256 \times 256$ patches. We have set the total training epochs to  500 and the mini-batch size to 8 across all datasets. The Adam optimizer \citep{Kingma2014Adam:Optimization} is used  with $\beta_1 = 0.9$, $\beta_2 = 0.999$, and $\epsilon = 1.0 \times 10^{-08}$ and an initial learning rate of $lr_{initial}=1e-4$, which linearly decays to $lr_{final}=1e-6$ in the final $200$ epochs.

\begin{figure*}[h]
\centering
\includegraphics[width=0.90\textwidth]{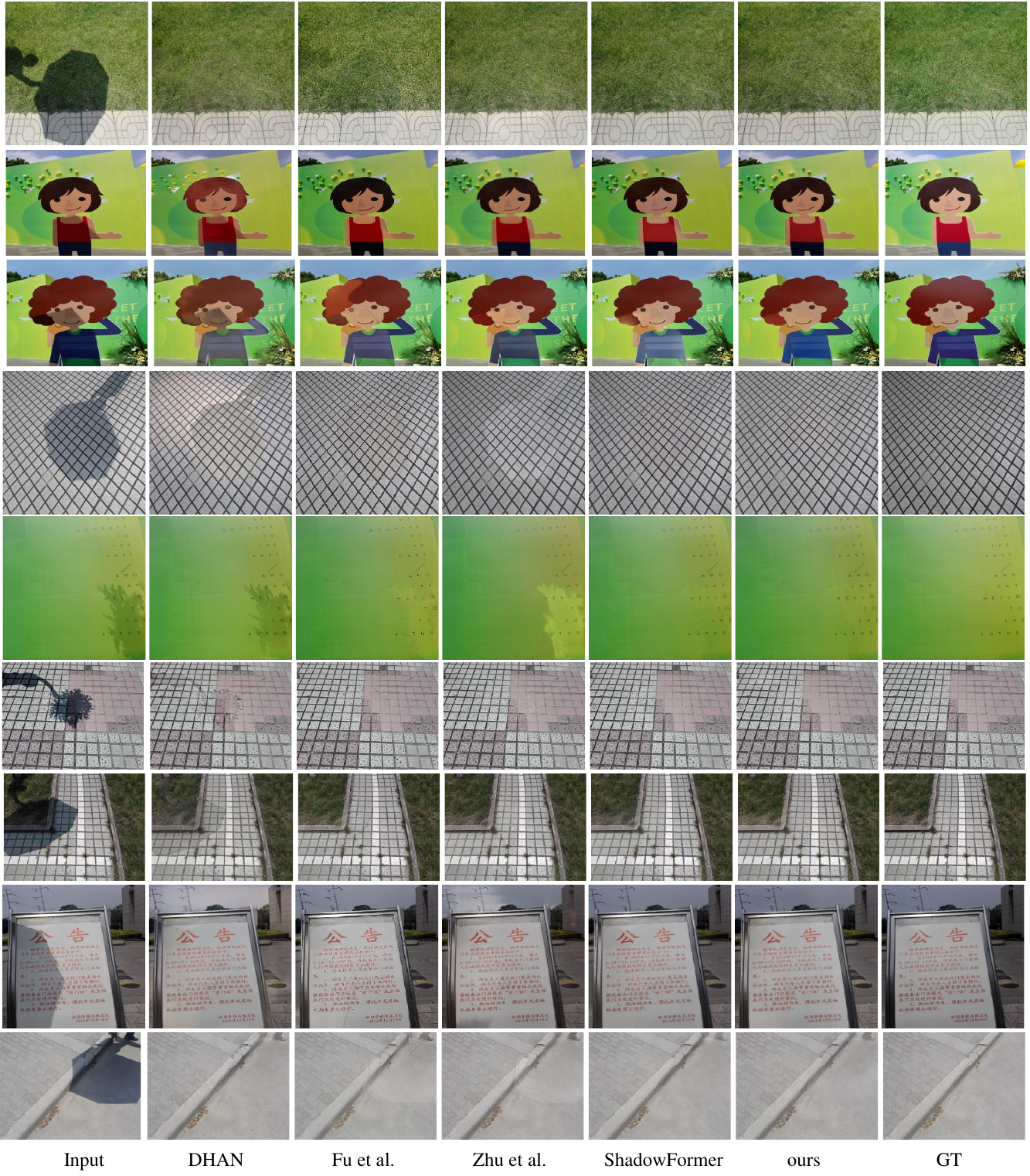}
\caption{A side-by-side comparison of our method with state-of-the-art shadow removal techniques on sample images from the ISTD dataset \citep{Wang2018StackedRemoval}. The images are arranged from left to right as follows: the input image, methods by DHAN \citep{Cun2020TowardsGAN.},  \citet{Fu2021Auto-exposureRemoval}, \citet{Zhu2022EfficientRemoval}, ShadowFormer \citep{Guo2023ShadowFormer:Removal}, our method, and the Ground Truth (GT). Zoom in to see the details.}
\label{fig:7}
\end{figure*}

\begin{figure*}[h]
\centering
\includegraphics[width=0.95\textwidth]{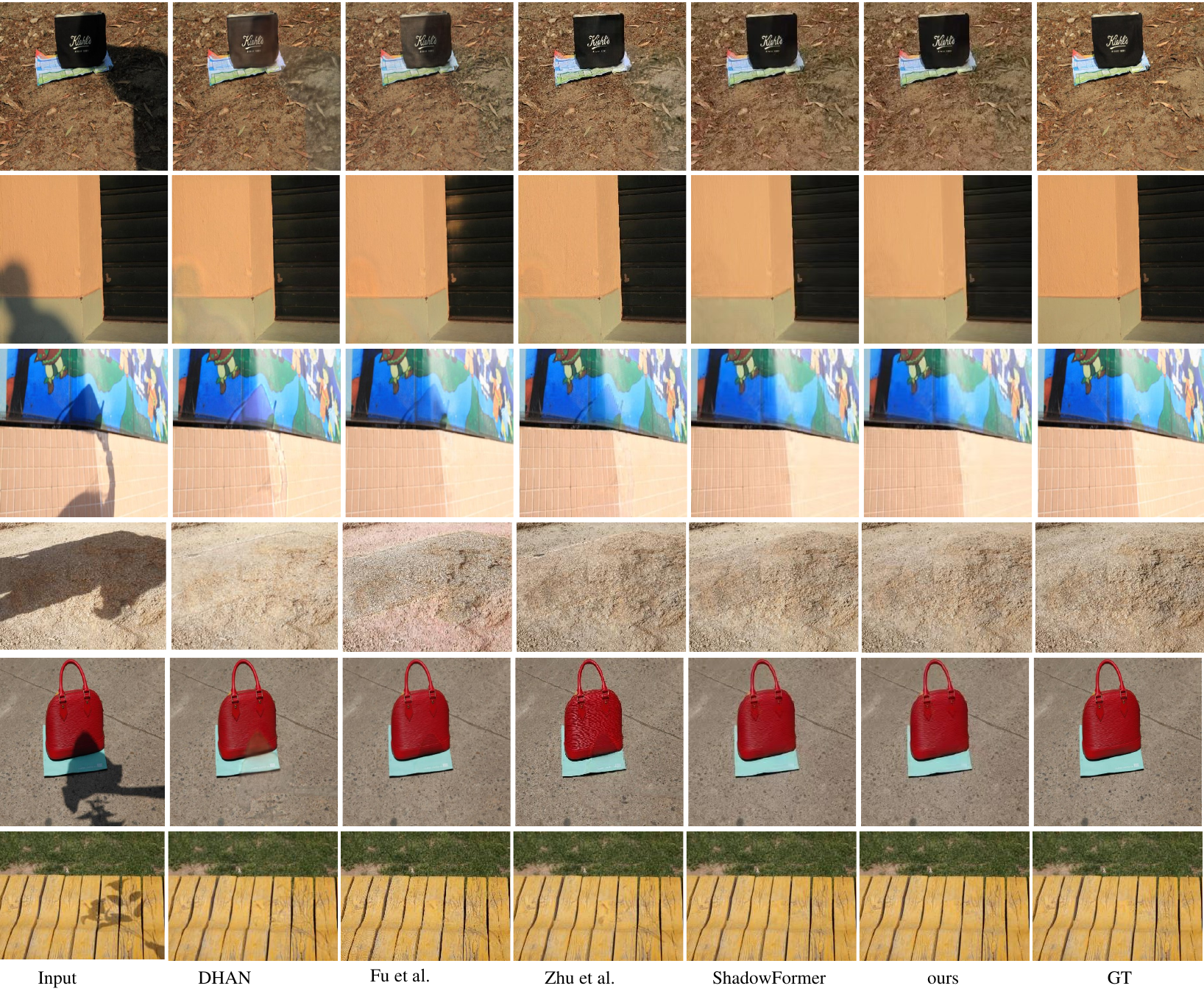}
\caption{A side-by-side comparison of our method with state-of-the-art shadow removal techniques on sample images from the SRD dataset \citep{Qu2017Deshadownet:Removal}. The images are arranged from left to right as follows: the input image, methods by  DHAN \citep{Cun2020TowardsGAN.},  \citet{Fu2021Auto-exposureRemoval}, \citet{Zhu2022EfficientRemoval}, ShadowFormer \citep{Guo2023ShadowFormer:Removal}, our method, and the Ground Truth (GT). Zoom in to see the details.}
\label{fig:8}
\end{figure*}

\begin{figure*}[h]
\centering
\includegraphics[width=0.95\textwidth]{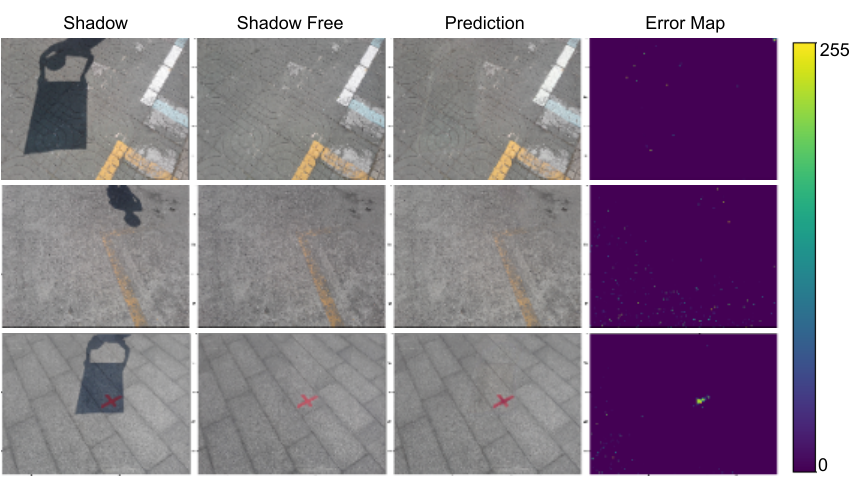}
\caption{Visual examples of error maps resulting from our method. The first column displays the original images with shadows. The second column shows the ground truth images with no shadows. The third column presents the predicted shadow-free images after achieved using FieldNet. The fourth column illustrates the error maps for the shadow removal.
The colour scale on the right side of the error maps ranges from 0 to 255, indicating the magnitude of the error.  Zoom in to see the details.}
\label{fig:5}
\end{figure*}

\begin{figure*}[ht]
\centering
\includegraphics[width=0.65\textwidth]{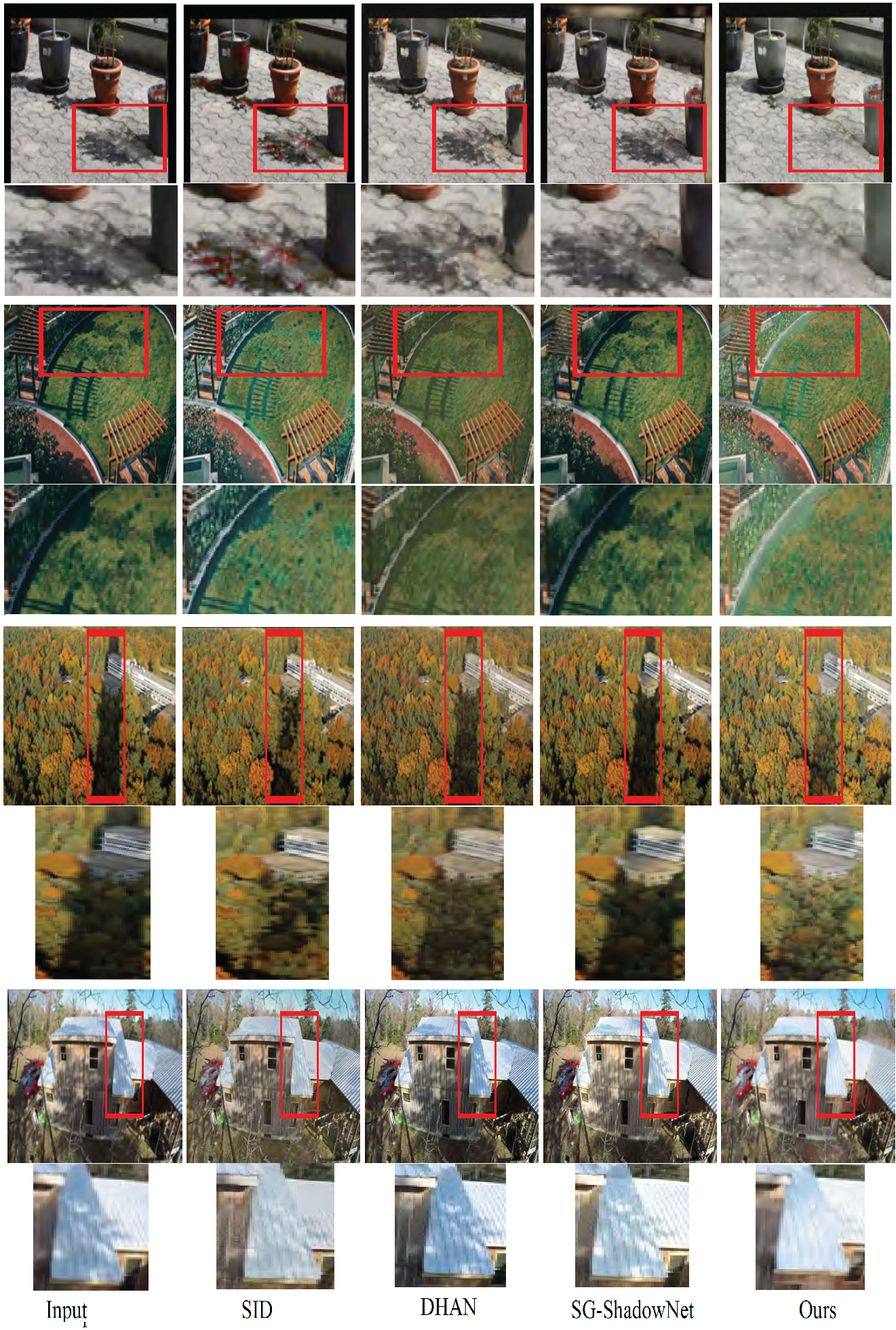}
\caption{A side-by-side comparison of our method with state-of-the-art shadow removal techniques on sample images from the SBU-Timelapse dataset \citep{Le2022Physics-BasedRemoval}. The images are arranged from left to right as follows: the input image, methods by SID \citep{Le2022Physics-BasedRemoval}, DHAN \citep{Cun2020TowardsGAN.},    SG-ShadowNet \citep{Wan2022Style-GuidedRemoval}, our method. It is important to note that the SBU-Timelapse dataset is an unpaired dataset, meaning it does not include shadow-free images as ground truth. This aspect adds an additional layer of complexity to the task of shadow removal and serves as a testament to the effectiveness of the compared methods. Zoom in to see the details.}
\label{fig:9}
\end{figure*}

\subsection{Comparison with State-of-the-Art Methods} \label{seccompare}

In this section, we provide a comparable evaluation of our proposed method with various state-of-the-art (SOTA) shadow removal algorithms. These include one traditional method, namely \citet{Guo2012PairedRemoval}, and a number of deep learning-based methods such as MaskShadow-GAN \citep{Hu2019Mask-ShadowGAN:Data}, ST-CGAN \citep{Wang2018StackedRemoval}, ARGAN \citep{Ding2019Argan:Removal}, DSC \citep{Hu2020Direction-AwareRemoval},  G2R \citep{Liu2021FromRemoval}, \citet{Fu2021Auto-exposureRemoval}, \citet{Jin2021DC-ShadowNet:Network}, DHAN \citep{Cun2020TowardsGAN.}, \citet{Zhu2022EfficientRemoval}, SG-ShadowNet \citep{Wan2022Style-GuidedRemoval}, BMNet \citep{Zhu2022BijectiveRemoval},  and ShadowFormer \citep{Guo2023ShadowFormer:Removal}.
Given the diversity in experimental settings adopted by prior shadow removal methods, we seek a fair comparison by conducting experiments on two primary settings over the ISTD dataset. These settings follow the methodologies of two recent works,  \citet{Zhu2022EfficientRemoval} and DHAN \citep{Cun2020TowardsGAN.}.
The first setting involves resizing the results into a resolution of $256\times 256$ for evaluation. The second setting maintains the original image resolutions throughout both the training and testing stages. 
It is important to note that all shadow removal results from the competing methods are either quoted from the original papers, taken from the previous papers or reproduced using their official implementations. This ensures the integrity and fairness of our comparative analysis.

\begin{table*}[h]
\caption{The quantitative results of shadow removal
using our models and recent state-of-the-art methods on ISTD~\citep{Wang2018StackedRemoval}
datasets.
We put ``-'' to denote unavailable data.}
\label{tab:istd_res}
\centering
\footnotesize
\setlength{\tabcolsep}{0.4em}
\renewcommand{\arraystretch}{1.0}
\vspace{-0.0cm}
\adjustbox{width=.95\linewidth}{
\begin{tabular}{c|l|l|l| ccc| ccc| ccc}
\toprule
\multirow{2}{*}{} &\multirow{2}{*}{Method} &\multirow{2}{*}{Params} & \multirow{2}{*}{Flops} & \multicolumn{3}{c|}{Shadow Region (S)}  &
\multicolumn{3}{c|}{Non-Shadow Region (NS)}  &
\multicolumn{3}{c}{All Image (ALL)} \\
& & & & PSNR$\uparrow$ & SSIM$\uparrow$ & RMSE$\downarrow$ & PSNR$\uparrow$ & SSIM$\uparrow$ & RMSE$\downarrow$ & PSNR$\uparrow$ & SSIM$\uparrow$ & RMSE$\downarrow$ \\
\midrule
\multirow{15}{*}{\rotatebox{90}{$256\times 256$}}  
&Input image                     &- &- &22.40 &0.936 &32.10 &27.32 &0.976 &7.09 &20.56 &0.893 &10.88 \\
&\citet{Guo2012PairedRemoval}               &- &- &27.76 &0.964 &18.65 &26.44 &0.975 &7.76 &23.08 &0.919 &9.26 \\
&ShadowGAN \citep{Hu2019Mask-ShadowGAN:Data}    &11.4M &56.8G &- &- &12.67 &- &- &6.68 &- &- &7.41 \\
&ST-CGAN \citep{Wang2018StackedRemoval}     & 29.2M & 17.9G & 33.74 & 0.981 & 9.99 & 29.51 & 0.958 & 6.05 & 27.44 & 0.929 & 6.65 \\
&ARGAN \citep{Ding2019Argan:Removal}       & 125.8M  & - & - & - & 6.65 & - & - & 5.41 & - & - & 5.89 \\
&DSC \citep{Hu2020Direction-AwareRemoval}          &  22.3M  &  123.5G &  34.64 &  0.984 &  8.72 &  31.26 &  0.969 &  5.04 &  29.00 &  0.944 &  5.59 \\
&DHAN \citep{Cun2020TowardsGAN.}          &  21.8M  &  262.9G &  35.53 &  0.988 &  7.73 &  31.05 &  0.971 &  5.29 &  29.11 &  0.954 &  5.66 \\
&G2R \citep{Liu2021FromRemoval}         &  22.8M  &  113.9G &  32.66 &  0.984 &  10.47 &  26.27 &  0.968 &  7.57 &  25.07 &  0.946 &  7.88 \\
& \citet{Fu2021Auto-exposureRemoval}               &  143.0M &  160.3G &  34.71  &  0.975 &  7.91 &  28.61 &  0.880 &  5.51 &  27.19 &  0.945 &  5.88 \\
& \citet{Jin2021DC-ShadowNet:Network}              &  21.2M  &  105.0G &  31.69   &  0.976 &  11.43 &  29.00 &  0.958 &  5.81 &  26.38 &  0.922 &  6.57 \\
& \citet{Zhu2022EfficientRemoval}             &  10.1M  &  56.1G  &36.95   &0.987 &8.29 &31.54 &0.978 &4.55 &29.85 &0.960 &5.09 \\
&BMNet \citep{Zhu2022BijectiveRemoval}        &  0.4M   &11.0G   &35.61 &0.988 &7.60 &32.80 &0.976 &4.59 &30.28 &0.959 &5.02 \\
&SG-ShadowNet \citep{Wan2022Style-GuidedRemoval} &  6.2M   &39.7G   &36.03 &0.988 &7.30 &32.56 &0.978 &4.38 &30.23 &0.961 &4.80 \\
&ShadowFormer \citep{Guo2023ShadowFormer:Removal} &  9.3M   &100.9G  &38.19 &0.991 &5.96 &34.32 &0.981 &3.72 &32.21 &0.968 &4.09 \\
&ShadowRefiner \citep{dong2024shadowrefiner}      &  15.3M   &120.3G  &37.85 &0.988 &5.90 &33.95 &0.978 &3.65 &28.75 &0.916 &4.57  \\
&FieldNet (Ours)                                             &  2.7M   &7.37G   &37.47 &0.990 &6.03 &33.80 &0.978 &4.06 &31.57 &0.965 &4.27\\
&FieldNet 'Plus' (Ours)                                       &  2.7M   &7.37G   &38.67 &0.991 &5.83 &34.40 &0.987 &3.54 &32.76 &0.967 &4.07\\
   
\midrule
\multirow{12}{*}{\rotatebox{90}{Original}}  
&Input image                    &-&-& 22.34 & 0.935 & 33.23 & 26.45 & 0.947 & 7.25 & 20.33 & 0.874 & 11.35 \\
&ARGAN \citep{Ding2019Argan:Removal}       &125.8M & -& -     & -     & 9.21  & -     & -     & 6.27 & -     & -     & 6.63  \\
&DSC \citep{Hu2020Direction-AwareRemoval}           & 22.3M  &  345.8G  & 33.45 & 0.967 & 9.76  & 28.18 & 0.885 & 6.14 & 26.62 & 0.845 & 6.67  \\
&DHAN \citep{Cun2020TowardsGAN.}        & 21.8M  &  710.43G  & 34.79 & 0.983 & 8.13  & 29.54 & 0.941 & 5.94 & 27.88 & 0.921 & 6.29  \\
&G2R \citep{Liu2021FromRemoval}          & 22.8M  &  296.14G  & 32.31 & 0.978 & 11.18 & 25.51 & 0.941 & 8.10 & 24.40 & 0.915 & 8.42  \\
& \citet{Fu2021Auto-exposureRemoval}              & 143.0M &  464.87G  & 33.59 & 0.958 & 8.73  & 27.01 & 0.794 & 6.24 & 25.71 & 0.745 & 6.62  \\
& \citet{Jin2021DC-ShadowNet:Network}              & 21.2M  &  294.0G  & 30.59 & 0.949 & 12.43 & 25.88 & 0.785 & 7.11 & 24.16 & 0.724 & 7.79  \\
& \citet{Zhu2022EfficientRemoval}            & 10.1M  &  151.47G  & 33.78 & 0.956 & 9.44  & 27.39 & 0.786 & 6.23 & 26.06 & 0.734 & 6.68  \\
&BMNet \citep{Zhu2022BijectiveRemoval}      & 0.4M   &  28.6G  & 34.84 & 0.983 & 8.31  & 31.14 & 0.949 & 5.16 & 29.02 & 0.929 & 5.59  \\
&SG-ShadowNet \citep{Wan2022Style-GuidedRemoval} & 6.2M   &  115.13G  & 35.17 & 0.982 & 8.21  & 30.86 & 0.950 & 5.04 & 28.95 & 0.928 & 5.48  \\
&ShadowFormer \citep{Guo2023ShadowFormer:Removal}& 9.3M   &  282.52G  & 37.03 & 0.985 & 6.76  & 32.20 & 0.953 & 4.44 & 30.47 & 0.935 & 4.79  \\
&ShadowRefiner \citep{dong2024shadowrefiner}      & 15.3M   & 270.43G  & 36.80 & 0.982 & 6.50  & 31.90 & 0.950 & 4.35 & 28.50 & 0.912 & 4.70 \\
&FieldNet (Ours)                                             & 2.7M   &  19.9G     & 36.65 & 0.985 & 7.02  & 31.15 & 0.949 & 4.72 & 29.76 & 0.923 & 5.08  \\
&FieldNet 'Plus' (Ours)                                      & 2.7M   &  19.9G     & 37.84 & 0.986 & 6.55  & 32.88 & 0.954 & 4.26 & 30.88 & 0.964 & 4.57  \\
    
\bottomrule
    \end{tabular}
}

\end{table*}

\color{blue}
\vspace{0.3cm}
\noindent\textbf{Quantitative Measure.}  
The quantitative results on the testing sets over ISTD, ISTD+, and SRD are presented in Tables (\cref{tab:istd_res}, \cref{tab:istdplus_res}, and \cref{tab:srd_res}), respectively, with generalization validated on SBU-Timelapse (\cref{tab:sbu}). FieldNet ‘Plus’ achieves superior performance across all datasets: ISTD (PSNR: 32.76, RMSE: 4.07), ISTD+ (RMSE: 2.6), SRD (SSIM: 0.981), and SBU-Timelapse (NRSS: 0.9632), outperforming ST-CGAN (ISTD PSNR: 27.44), ARGAN (ISTD RMSE: 5.89), and ShadowFormer (ISTD PSNR: 32.21). Compared to these methods, FieldNet’s lightweight design (2.7M parameters vs. 29.2M for ST-CGAN, 125.8M for ARGAN) ensures efficiency without sacrificing quality. Its unpaired training approach and large synthetic dataset enhance robustness across diverse scenes, unlike paired-data-dependent methods.
\color{black}


\color{blue}
\vspace{0.3cm}
\noindent\textbf{Qualitative Measure.}  
Qualitative results on ISTD (\cref{fig:7}) and SRD (\cref{fig:8}) showcase FieldNet’s advantages over ST-CGAN, ARGAN, and ShadowFormer. ST-CGAN and ARGAN often produce illumination mismatches and boundary artifacts (e.g., fourth row, \cref{fig:7}), while FieldNet maintains tonal consistency. ShadowFormer excels in fine detail preservation (e.g., fifth row, \cref{fig:7}), but FieldNet avoids sharp boundaries and ghosting seen in ST-CGAN and ARGAN, as evidenced in \cref{fig:8}’s third row with complex structures. This balance of quality and efficiency highlights FieldNet’s suitability for real-time applications.
\color{black}

The ISTD examples contain relatively simple scenes with high contextual similarity across images. Despite this, prior arts often generate illumination mismatches and artifacts along shadow boundaries in these samples.
By contrast, our method produces results absent of tonal inconsistencies and edge effects. This holds even for subtle shadows in similar contexts that cause issues for existing techniques. We also demonstrate strong generalizability on the more complex and diverse SRD shadows. The ability to attenuate difficult shadows while preserving lighting coherence emphasizes the robustness of our data-driven framework.
For instance,  \citet{Fu2021Auto-exposureRemoval} and DHAN \citep{Cun2020TowardsGAN.} methods erroneously brighten some areas with poor lightness, such as the fourth row of \cref{fig:7}, resulting in ghosting artifacts.
Furthermore, almost all competing methods except ShadowFormer \citep{Guo2023ShadowFormer:Removal} fail to maintain the illumination coherencies among shadow and non-shadow areas, which severely disrupts the image structure and patterns, thereby generating sharp boundaries as shown in the fourth and fifth rows of \cref{fig:7}.
However, with the benefits of the proposed mask dissociation loss, It is evident that our techniques can effectively balance the supervision between shadow boundary pixels and pixels inside shadows. This improves shadow removal, even in complex lighting conditions and intricate shadow patterns. Furthermore, our approach enhances the overall image quality by preserving the integrity of non-shadow regions, thereby providing a more natural and visually pleasing result.
On the contrary, the image structures on the SRD dataset are more complex and often feature various colours. In these SRD dataset samples, previous work has sometimes struggled to accurately remove shadows, particularly in areas with intricate structures or varied colours, such as the blue poster in the third row of \cref{fig:8}. The complexity of the image structures and the diversity of colours can pose significant challenges for shadow removal algorithms, leading to less-than-optimal results. Despite the complexity and colour diversity, our method effectively removes shadows while preserving the original colours and structures. This results in high-quality images that maintain the integrity of the original scene, proving the effectiveness of our approach in handling complex and diverse image data.

\vspace{0.3cm}
\noindent\textbf{Error Visualization:} To further analyze the performance of our method, we present error maps in Fig.~\ref{fig:5}. These error maps provide a visual representation of the difference between the ground truth (shadow-free image) and the output of FieldNet. The first column of Fig.~\ref{fig:5} displays the original images with shadows. The second column shows the ground truth images with no shadow. The third column presents the predicted shadow-free images after our shadow-removal process. The most important information lies in the fourth column, which illustrates the error maps for the predictions. Coloured areas in these error maps represent the pixels where the largest discrepancies between the ground truth and the prediction occur. The colour scale on the right side of the error maps ranges from 0 to 255, with higher values indicating a larger error (i.e., a greater difference between the predicted shadow-free image and the ground truth). By examining these error maps, we can gain valuable insights into the strengths and weaknesses of our approach.

\begin{table*}[h]
\caption{The quantitative results of shadow removal
using our models and recent state-of-the-art methods on SRD~\citep{Qu2017Deshadownet:Removal}
datasets.
We put ``-'' to denote unavailable data.}
\label{tab:srd_res}
\centering
\footnotesize
\setlength{\tabcolsep}{0.4em}
\renewcommand{\arraystretch}{1.0}
\vspace{-0.0cm}
\adjustbox{width=.95\linewidth}{
\begin{tabular}{c|l|  ccc| ccc| ccc}
\toprule
\multirow{2}{*}{} &\multirow{2}{*}{Method}   & \multicolumn{3}{c|}{Shadow Region (S)}  &
\multicolumn{3}{c|}{Non-Shadow Region (NS)}  &
\multicolumn{3}{c}{All Image (ALL)} \\
  & & PSNR$\uparrow$ & SSIM$\uparrow$ & RMSE$\downarrow$ & PSNR$\uparrow$ & SSIM$\uparrow$ & RMSE$\downarrow$ & PSNR$\uparrow$ & SSIM$\uparrow$ & RMSE$\downarrow$ \\
\midrule
\multirow{13}{*}{\rotatebox{90}{$256\times 256$}}  
&Input image                     & 18.96 & 0.871 & 36.69 & 31.47 & 0.975 & 4.83 & 18.19 & 0.830 & 14.05 \\
& \citep{Guo2012PairedRemoval}               & -     & -     & 29.89 & -     & -     & 6.47 & -     & -     & 12.60 \\
&DeShadowNet \citep{Qu2017Deshadownet:Removal}   & -     & -     & 11.78 & -     & -     & 4.84 & -     & -     & 6.64  \\
&DSC \citep{Hu2020Direction-AwareRemoval}            & 30.65 & 0.960 & 8.62  & 31.94 & 0.965 & 4.41 & 27.76 & 0.903 & 5.71  \\
&ARGAN \citep{Ding2019Argan:Removal}        & -     & -     & 6.35  & -     & -     & 4.46 & -     & -     & 5.31  \\
&DHAN \citep{Cun2020TowardsGAN.}          & 33.67 & 0.978 & 8.94  & 34.79 & 0.979 & 4.80 & 30.51 & 0.949 & 5.67  \\
& \citet{Fu2021Auto-exposureRemoval}                & 32.26 & 0.966 & 8.55  & 31.87 & 0.945 & 5.74 & 28.40 & 0.893 & 6.50  \\
& \citet{Jin2021DC-ShadowNet:Network}               & 34.00 & 0.975 & 7.70  & 35.53 & 0.981 & 3.65 & 31.53 & 0.955 & 4.65  \\
& \citet{Zhu2022EfficientRemoval}             & 34.94 & 0.980 & 7.44  & 35.85 & 0.982 & 3.74 & 31.72 & 0.952 & 4.79  \\
&BMNet \citep{Zhu2022BijectiveRemoval}        & 35.05 & 0.981 & 6.61  & 36.02 & 0.982 & 3.61 & 31.69 & 0.956 & 4.46  \\
&SG-ShadowNet \citep{Wan2022Style-GuidedRemoval} & 36.55 & 0.981 & 7.56  & 34.23 & 0.961 & 3.06 & 31.31 & 0.927 & 4.30  \\
&ShadowFormer \citep{Guo2023ShadowFormer:Removal} & 36.91 & 0.989 & 5.90  & 36.22 & 0.989 & 3.44 & 32.90 & 0.958 & 4.04  \\
&ShadowRefiner \citep{dong2024shadowrefiner}      & 36.50 & 0.986 & 5.70  & 35.80 & 0.985 & 3.30 & 32.50 & 0.950 & 3.95  \\
&FieldNet (Ours)                                              & 35.89 & 0.974 & 5.43  & 36.01 & 0.983 & 3.57 & 32.03 & 0.940 & 4.14  \\
&FieldNet 'Plus' (Ours)                                       & 37.89 & 0.990 & 4.86  & 37.21 & 0.993 & 3.33 & 33.17 & 0.981 & 3.56  \\
\midrule
\multirow{8}{*}{\rotatebox{90}{Original}}  
&Input image               & 19.00 & 0.871 & 39.23 & 28.41 & 0.949 & 5.86  & 17.87 & 0.804 & 14.62 \\
&DSC \citep{Hu2020Direction-AwareRemoval}      & 25.95 & 0.912 & 20.40 & 22.46 & 0.748 & 16.89 & 20.15 & 0.642 & 17.75 \\
&DHAN \citep{Cun2020TowardsGAN.}    & 32.21 & 0.969 & 8.39  & 30.58 & 0.943 & 5.02  & 27.70 & 0.898 & 5.88  \\
& \citet{Fu2021Auto-exposureRemoval}         & 31.19 & 0.955 & 9.65  & 28.10 & 0.894 & 6.63  & 25.83 & 0.825 & 7.32  \\
& \citet{Jin2021DC-ShadowNet:Network}         & 31.21 & 0.955 & 9.23  & 28.62 & 0.896 & 5.91  & 26.18 & 0.827 & 6.72  \\
& \citet{Zhu2022EfficientRemoval}        & 28.25 & 0.930 & 11.57 & 23.83 & 0.803 & 8.48  & 21.95 & 0.700 & 9.19  \\
&BMNet \citep{Zhu2022BijectiveRemoval}  & 33.28 & 0.973 & 7.84  & 32.71 & 0.963 & 4.38  & 29.21 & 0.923 & 5.24  \\
&FieldNet (Ours)                                    & 33.08 & 0.955 & 7.95  & 31.77 & 0.942 & 4.86  & 28.66 & 0.905 & 5.85  \\
&FieldNet 'Plus' (Ours)                        & 34.12 & 0.981 & 6.77  & 33.45 & 0.978 & 3.98  & 30.43 & 0.935 & 4.76  \\
    
\bottomrule
    \end{tabular}
}

\end{table*}

\begin{table*}[h]
\caption{The quantitative results of shadow removal
using our models and recent state-of-the-art methods on ISTD+
 datasets (Le and Samaras, 2019).}
\label{tab:istdplus_res}
\centering
\footnotesize
\setlength{\tabcolsep}{0.4em}
\renewcommand{\arraystretch}{1.0}
\adjustbox{width=.95\linewidth}{
\begin{tabular}{c|l|ccc }
\toprule
\multirow{2}{*}{} &\multirow{2}{*}{Method} & \multicolumn{3}{c}{RMSE$\downarrow$} \\

& & Shadow Region (S) & Non-Shadow Region (NS) & All Image (ALL) \\

\midrule
\multirow{14}{*}{\rotatebox{90}{$256\times 256$}}  
&Input images                          & 39.0 & 2.6 & 8.4 \\
&  \citep{Guo2012PairedRemoval}                     & 22.0 & 3.1 & 6.1 \\
&ST-CGAN \citep{Wang2018StackedRemoval}           & 13.4 & 7.7 & 8.7 \\
&DeshadowNet \citep{Qu2017Deshadownet:Removal}        & 15.9 & 6.0 & 7.6 \\
&Mask-ShadowGAN \citep{Hu2019Mask-ShadowGAN:Data}     & 12.4 & 4.0 & 5.3 \\
&Param+M+D-Net \citep{Le2020FromRemoval}  & 9.7 & 3.0 & 4.0 \\
&G2R \citep{Liu2021FromRemoval}                 & 7.3  & 2.9 & 3.6 \\
&SP+M-Net \citep{Le2019ShadowDecomposition}      & 7.9 & 3.1 & 3.9 \\
& \citet{Fu2021Auto-exposureRemoval}                     & 6.7  & 3.8 & 4.2 \\
& \citet{Jin2021DC-ShadowNet:Network}                     & 10.4 & 3.6 & 4.7 \\
&SG-ShadowNet \citep{Wan2022Style-GuidedRemoval}       & 5.9  & 2.9 & 3.4 \\
&BMNet \citep{Zhu2022BijectiveRemoval}              & 5.6  & 2.5 & 3.0 \\
&ShadowFormer \citep{Guo2023ShadowFormer:Removal} & \textbf{5.2} & \textbf{2.3} & \textbf{2.8} \\ &FieldNet (Ours)  & 5.7 & 2.8 & 3.4 \\ &FieldNet 'Plus' (Ours)  & \textbf{4.9} & \textbf{2.1} & \textbf{2.6}\\
 \midrule
\multirow{7}{*}{\rotatebox{90}{Original}}  
&Input images                     & 38.5 & 3.3 & 9.2 \\
& \citet{Fu2021Auto-exposureRemoval}               & 10.4 & 8.0 & 8.4 \\
& \citet{Jin2021DC-ShadowNet:Network}               & 11.9 & 5.3 & 6.3 \\
&SG-ShadowNet \citep{Wan2022Style-GuidedRemoval} & 7.3  & 3.8 & 4.3 \\
&BMNet \citep{Zhu2022BijectiveRemoval}        & 6.6  & 3.2 & 3.7 \\
&ShadowFormer \citep{Guo2023ShadowFormer:Removal} & \textbf{6.2} & \textbf{3.2} & \textbf{3.6} \\ &FieldNet (Ours)  & 6.7 & 3.6 & 3.9 \\ &FieldNet 'Plus' (Ours)  & \textbf{5.9} & \textbf{3.0} & \textbf{3.2} \\
\bottomrule
    \end{tabular}
}

\end{table*}

\subsection{Generalization Ability}  
To validate the generalization ability of our proposed approach, we conducted experiments on the SBU-Timelapse dataset \citep{Le2022Physics-BasedRemoval}. This dataset is unpaired, meaning it does not include shadow-free images as ground truth. Therefore, we adopted non-reference evaluation metrics: \textbf{Gradient Structure Similarity (NRSS)} and \textbf{Natural Image Quality Evaluator (NIQE)} \citep{Li2022APEquation}.  

\begin{itemize}
    \item \textbf{NRSS}: Measures the naturalism of the generated shadow-free images. A higher NRSS value indicates better performance, meaning the generated image is more similar to a natural image.  
    \item \textbf{NIQE}: Evaluates perceptual quality. A lower NIQE value indicates better perceptual quality, meaning the image appears more natural to human observers.  
\end{itemize}

We compared our method with several state-of-the-art methods that do not require masks during inference, including \textbf{SID} \citep{Le2022Physics-BasedRemoval}, \textbf{DHAN} \citep{Cun2020TowardsGAN.}, and \textbf{SG-ShadowNet} \citep{Wan2022Style-GuidedRemoval}. As shown in Table \ref{tab:sbu}, our approach surpasses the other methods across all metrics. Specifically:  
\begin{itemize}
    \item Compared to \textbf{SG-ShadowNet}, our method reduced the NIQE from \textbf{7.6532} to \textbf{3.7532} and increased the NRSS from \textbf{0.9324} to \textbf{0.9632}.  
    \item Our method also outperformed \textbf{SID} and \textbf{DHAN}, demonstrating superior generalization ability in handling diverse and complex image data.  
\end{itemize}

Figure \ref{fig:9} shows that our method achieves reasonable perceptual performance in challenging environments. While some methods, such as ShadowFormer in Figures \ref{fig:7} and \ref{fig:8}, may produce visually detailed results that appear closer to the ground truth in certain cases, our method emphasizes \textbf{global illumination consistency}, which is particularly effective in complex lighting conditions. This trade-off allows our method to achieve superior quantitative performance while maintaining high perceptual quality, as evidenced by the NRSS and NIQE metrics.  

These results highlight the effectiveness of our proposed method in handling complex and diverse image data, demonstrating its strong generalization ability. In future work, we aim to improve the preservation of fine-grained details further while maintaining global illumination consistency.

\begin{table*}[h]
\caption{Quantitative comparison of our method with the state-of-the-art methods on the SBU-Timelapse dataset~\citep{Le2022Physics-BasedRemoval}. 
``$\uparrow$'' indicates the higher the better and ``$\downarrow$'' indicates the lower the better.}
\label{tab:sbu}
\centering
\footnotesize
\setlength{\tabcolsep}{0.4em}
\renewcommand{\arraystretch}{1.0}
\vspace{-0.0cm}
\adjustbox{width=.50\linewidth}{
\begin{tabular}{l| cc }
\toprule
\multirow{2}{*}{Method}  &  
\multicolumn{2}{c}{All Image (ALL)} \\
  & NRSS$\uparrow$ & NIQE$\downarrow$  \\
\midrule
SID~\citep{Le2022Physics-BasedRemoval}               &0.9432 &5.1843 \\
DHAN   \citep{Cun2020TowardsGAN.}                    &0.9365 &9.6424 \\
SG-ShadowNet    \citep{Wan2022Style-GuidedRemoval}   &0.9324 &7.6532 \\
\textbf{FieldNet (Ours) }                                                 &\textbf{0.9632} &\textbf{3.7532}\\
\bottomrule
    \end{tabular}
}
\end{table*}

\color{blue}
To further enhance FieldNet’s generalization across datasets like ISTD, ISTD+, SRD, and SBU-Timelapse, modifications could include integrating real-world paired shadow/shadow-free data to reduce synthetic bias, applying multi-domain augmentations (e.g., urban, indoor scenes) to broaden scene diversity, and implementing adaptive loss weighting to handle extreme lighting conditions. These enhancements, detailed in \cref{secdisc}, aim to address limitations in complex shadow patterns and improve robustness beyond the current synthetic dataset’s scope.
\color{black}



\color{blue}
\subsection{Computational efficiency}  
In this section, we analyse the computational efficiency of our proposed shadow removal method in comparison with several state-of-the-art models. FieldNet achieves 66 FPS on an Nvidia 2080Ti through its lightweight U-Net-based architecture (2.7M parameters, 7.37G FLOPs) and single-pass inference, avoiding the iterative processes of GANs (e.g., DHAN: 9 FPS, 262.9G FLOPs) or diffusion models (e.g., Mei et al., 2024: ~0.5-1 FPS). Compared to ShadowFormer (8 FPS, 100.9G FLOPs) and ST-CGAN (slow due to 29.2M parameters), FieldNet offers 7-9x higher speed with significantly lower computational demands, as shown in \cref{tab:flops} and \cref{fig:1}, making it highly practical for real-time field robotics.
\color{black}

\begin{table*}[h]
\caption{A comparison of our method with state-of-the-art models on key performance and complexity analysis metrics including PSNR, inference time, frames per second (FPS), number of parameters (Params), and computational complexity (gigaFLOPs). All models were tested on the ISTD dataset using a 2080Ti GPU device.}
\label{tab:flops}
\centering
\footnotesize
\setlength{\tabcolsep}{0.4em}
\renewcommand{\arraystretch}{1.0}
\vspace{-0.0cm}
\adjustbox{width=.90\linewidth}{
\begin{tabular}{l|c|c|c|c|c}
\toprule
 {Method} &  {FPS} &  {Time} &  {FLOPs} &  {PSNR} &  {Params}  \\
\midrule

DHAN \citep{Cun2020TowardsGAN.} & 9 & 0.117 & 262.9 & 29.12 & 21.8 \\
SP+M-Net \citep{Le2019ShadowDecomposition} & 10 & 0.105 & 160.1 & 25.08 & 141.2 \\
G2R \citep{Liu2021FromRemoval}    & 4 & 0.254 & 113.9 & 25.07 & 113.9 \\
\citet{Fu2021Auto-exposureRemoval}  & 7 & 0.155 & 160.3 & 27.19 & 143.0 \\
BMnet \citep{Guo2023ShadowFormer:Removal}   & 16 & 0.062 & 11.0 & 30.28 & \textbf{0.4} \\
ShadowFormer \citep{Guo2023ShadowFormer:Removal}   & 8 & 0.127 & 100.9 & 32.21 & 9.3 \\
\textbf{FieldNet (Ours) } & \textbf{66} & \textbf{0.015} & \textbf{7.5} & \textbf{32.76} & {2.7} \\

\bottomrule
    \end{tabular}
}
\end{table*}

\begin{figure*}[h]
\centering
\includegraphics[width=0.95\textwidth]{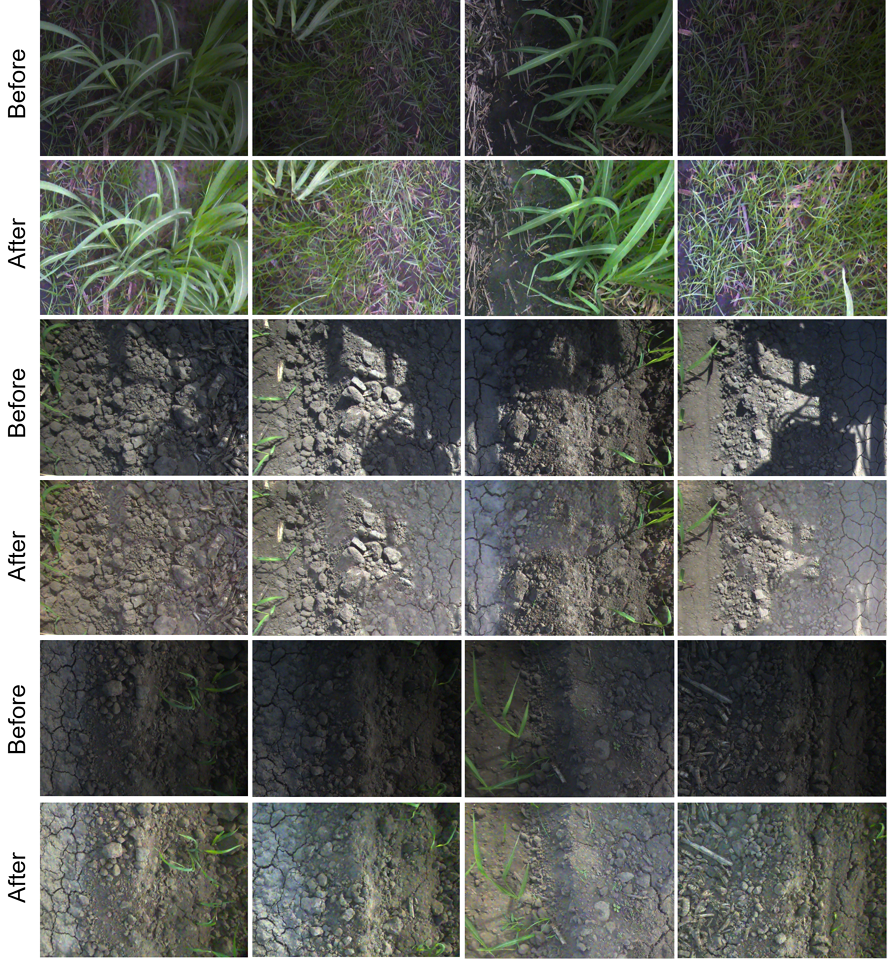}
\caption{Visual Enhancement in Robotic Weed Spot-Spraying: A Comparative Study of Images Before and After Shadow Removal}
\label{fig:4}
\end{figure*}

\subsection{Real-world case study of FieldNet in field robotics}
In this section, we present the results of applying ShaodowRemovalNet to a real-world field robotic setting. We show that
FieldNet presents substantial benefits for two critical areas of field robotics: real-time shadow removal and image enhancement during dataset labelling.

\subsubsection{Real-time shadow removal in agriculture field robotic}
Field robots often operate in dynamic environments with varying lighting conditions. This can significantly impact the uniformity of captured images, hindering the performance of computer vision pipelines within the robot. FieldNet addresses this challenge by enabling real-time image processing. By removing shadows from the captured frames, the method ensures consistent image quality and improves the clarity of crucial features for object detection and recognition tasks. This is particularly valuable in agricultural settings where environmental lighting fluctuates throughout the day. Consistent image quality allows the robot's computer vision system to perform more accurate object detection and recognition tasks, leading to improved decision-making and overall performance.

Figure~\ref{fig:4} presents real-world images captured by a weed control robot \citep{RahimiAzghadi2024}, before and after shadow removal. The clear improvement in image quality and visibility of the weeds demonstrates the effectiveness of our method in removing shadows.

In addition, Tables~\ref{tab:trial_data1} and~\ref{tab:trial_data2} present the measured inference times (ms) and frame rates (fps) of the shadow algorithm, respectively. These tables quantify the performance of the algorithm on a Jetson-Orin AGX platform in various configurations. They explore the impact of input size (256x256 and 512x512 pixels) and weight precision (float32, float16, int8) for both PyTorch and TensorRT implementations. Analyzing these results is crucial for selecting the optimal configuration for real-world applications, striking a balance between speed and accuracy requirements.

\subsubsection{Image enhancement for improved dataset labelling}
Training robust and accurate deep learning models is highly dependent on the quality of the labelled datasets used. Human experts typically label these datasets by manually identifying and annotating objects of interest within the images. Our method can be applied during this labelling process to preprocess the images before they are presented to human labellers. By removing shadows, the method enhances the visibility of key features within the image. This reduces the potential for human error during labeling and ultimately leads to a higher quality labeled dataset. A well-labeled dataset is essential for training deep learning models that can generalize well to unseen data and perform effectively in real-world scenarios.

As shown in Figure~\ref{fig:4}, the image quality and visibility of the weeds achieved after applying FieldNet signifies the effectiveness of our method in improving the labelling process, which could lead to improved deep learning of weed images.

\begin{figure*}[h]
\centering
\includegraphics[width=0.99\textwidth]{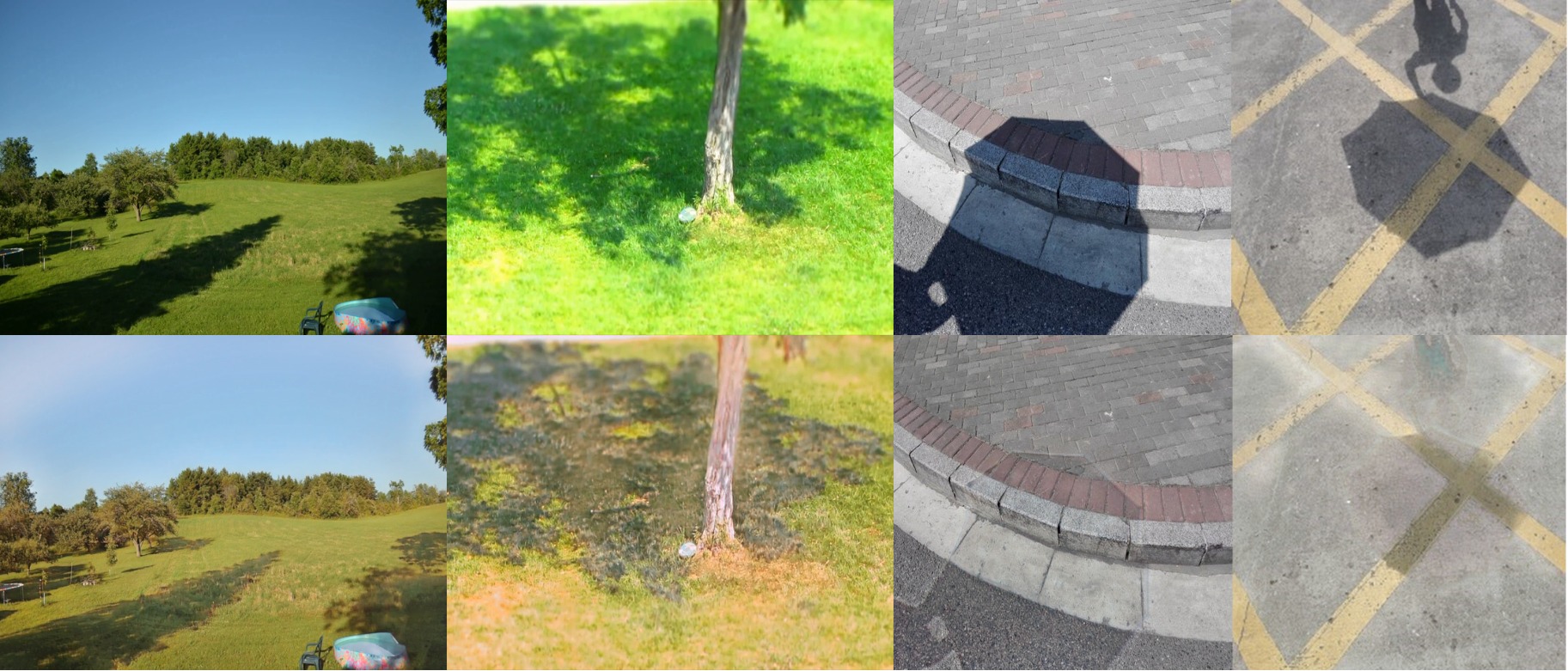}
\caption{Qualitative analysis of FieldNet's sub-optimal performance. Examples include (top row) high-contrast shadow edges and uneven illumination and (bottom row) complex textures where residual artefacts remain after shadow removal. These edge cases highlight the challenges of achieving consistent shadow-free images in diverse scenarios.}
\label{fig:10}
\end{figure*}

\begin{table}[h]
    \begin{center}
        \caption{Inference times (ms) of the shadow algorithm measured on a Jetson Orin AGX embedded GPU platform varying input size (512$\times$512 and 256$\times$256 px) and weight precision (float32, float16, int8) for PyTorch and TensorRT implementations.}
        \begin{tabular}{|c|c|c|c|}
        \hline
        \multirow{2}{*}{Environment} & \multirow{2}{*}{Weight Precision} & \multicolumn{2}{c|}{Input Size} \\  
        & & 512$\times$512 px & 256$\times$256 px \\
        \hline
        PyTorch & float32 & 80.87 $\pm$ 2.08 & 22.07 $\pm$ 0.42 \\
        TensorRT & float32 & 78.32 $\pm$ 0.25 & 21.52 $\pm$ 0.36 \\
        TensorRT & float16 & 51.79 $\pm$ 0.26 & 20.25 $\pm$ 0.26 \\
        TensorRT & int8 & 46.96 $\pm$ 0.34 & 18.80 $\pm$ 0.24 \\
        \hline
        \end{tabular}  
    \end{center}

    \label{tab:trial_data1}
\end{table}

\begin{table}[h]
    \centering
    \begin{center}
        \caption{Frame rates (fps) of the shadow algorithm measured on a Jetson Orin AGX embedded GPU platform varying input size (512$\times$512 and 256$\times$256 px) and weight precision (float32, float16, int8) for PyTorch and TensorRT implementations.}
        \begin{tabular}{|c|c|c|c|}
        \hline
        \multirow{2}{*}{Environment} & \multirow{2}{*}{Weight Precision} & \multicolumn{2}{c|}{Input Size} \\ 
        & & 512$\times$512 px & 256$\times$256 px \\
        \hline
        PyTorch & float32 & 12.37 $\pm$ 0.31 & 45.33 $\pm$ 0.84 \\
        TensorRT & float32 & 12.77 $\pm$ 0.04 & 46.47 $\pm$ 0.77 \\
        TensorRT & float16 & 19.31 $\pm$ 0.10 & 49.39 $\pm$ 0.63 \\
        TensorRT & int8 & 21.30 $\pm$ 0.15 & 53.21 $\pm$ 0.65 \\
        \hline
        \end{tabular}  
    \end{center}

    \label{tab:trial_data2}
\end{table}


\color{blue}
\subsection{Qualitative Analysis of Sub-Optimal Performance}  
While FieldNet demonstrates robust shadow removal in most scenarios, there are edge cases where its performance is less effective. Figure~\ref{fig:10} illustrates such examples, including high-contrast shadow edges, complex textures, and uneven illumination. In cases with high-contrast edges (top row), residual artifacts persist due to challenges in precisely delineating sharp boundaries, reducing PSNR by approximately 1-2 dB in these regions (\cref{tab:ablation_study}). Complex textures (bottom row) confuse feature modulation, leading to inconsistent color correction and minor artifacts. These limitations suggest areas for future improvement, such as refining the Probabilistic Enhancement Module (PEM) to better handle intricate shadow patterns or incorporating domain-specific augmentations during training to enhance texture robustness, as discussed in \cref{secdisc}.
\color{black}

\subsection{Ablation Study}  
An ablation study was conducted to understand the contribution of each component of the loss function and the Physics-Guided Enhancement Module (PEM) to the performance of the FieldNet model. The loss function is composed of three main components: the enhancement loss \(L_e\), the KL divergence \((L_m + L_s)\), and the boundary loss \(L_b\). Each of these components is weighted by a corresponding weight parameter (\(\alpha\), \(\beta\), and \(\gamma\) respectively) in the total loss function presented in Eq. \ref{equ13}.

To evaluate the impact of PEM, we generated 10,000 synthetic samples using the shadow synthesis process described in Section 3.2. These samples were used to train and validate the model, ensuring a robust evaluation of PEM's contribution. The results, as shown in \cref{tab:ablation_study}, demonstrate that PEM significantly improves the model's ability to handle complex lighting conditions and generate high-quality shadow-free images.

In the ablation study, each component of the loss function was removed one at a time, and the performance of the model was evaluated without that component. This allowed us to quantify the impact of each component on the overall performance of the model. The results of the ablation study, as shown in \cref{tab:ablation_study}, revealed the following:
\begin{itemize}
    \item The enhancement loss \(L_e\) was found to be crucial for generating high-quality shadow-free images, as it directly optimizes the pixel-wise reconstruction accuracy.
    \item The KL divergence \((L_m + L_s)\) played a key role in aligning the posterior and prior distributions in the latent space, enabling the model to incorporate diversity into the shadow removal process.
    \item The boundary loss \(L_b\) was particularly important for enhancing the supervision of pixels near the shadow boundaries, leading to more accurate shadow removal in these areas.
\end{itemize}

\begin{table*}[h]
\caption{Contributions of the components of the loss function and PEM to the performance of the FieldNet model tested on the ISTD~\citep{Wang2018StackedRemoval} dataset.}
\label{tab:ablation_study}
\centering
\footnotesize
\setlength{\tabcolsep}{0.4em}
\renewcommand{\arraystretch}{1.0}
\vspace{-0.0cm}
\adjustbox{width=.95\linewidth}{
\begin{tabular}{l|ccc|ccc|ccc}
\toprule
\multirow{2}{*}{Component}   & \multicolumn{3}{c|}{Shadow Region (S)} & \multicolumn{3}{c|}{Non-Shadow Region (NS)} & \multicolumn{3}{c}{All Image (ALL)} \\
 & PSNR\(\uparrow\) & SSIM\(\uparrow\) & RMSE\(\downarrow\) & PSNR\(\uparrow\) & SSIM\(\uparrow\) & RMSE\(\downarrow\) & PSNR\(\uparrow\) & SSIM\(\uparrow\) & RMSE\(\downarrow\) \\
\midrule
\(L_e\)               & 19.12 & 0.672 & 9.85 & 20.45 & 0.742 & 9.12 & 20.76 & 0.752 & 8.36 \\
\(L_m + L_s\)         & 36.54 & 0.980 & 6.92 & 32.12 & 0.948 & 4.78 & 30.45 & 0.960 & 4.89 \\
\(L_b\)               & 35.12 & 0.944 & 7.72 & 32.04 & 0.950 & 4.64 & 28.88 & 0.864 & 5.89 \\
PEM                  & 36.88 & 0.978 & 6.45 & 32.45 & 0.952 & 4.32 & 30.12 & 0.962 & 4.65 \\
Total Loss \((L)\)    & \textbf{37.84} & \textbf{0.986} & \textbf{6.55} & \textbf{32.88} & \textbf{0.954} & \textbf{4.26} & \textbf{30.88} & \textbf{0.964} & \textbf{4.57} \\
\bottomrule
\end{tabular}
}
\end{table*}


\color{blue}
The ablation study highlights the importance of each component in the loss function and the significant contribution of PEM to the overall performance of FieldNet. Specifically, the novel loss function increases PSNR by 1.3 dB (from 36.54 to 37.84) by improving boundary accuracy through edge-aware supervision. Mask dissociation enhances SSIM by 0.02 (from 0.944 to 0.986) by refining shadow edges via body and detail mask separation. PEM boosts PSNR by 0.8 dB (from 36.88 to 37.84) by diversifying shadow-free outputs, enhancing robustness to lighting variations. Together, these components elevate FieldNet’s performance to a PSNR of 30.88, SSIM of 0.964, and RMSE of 4.57 on ISTD, far surpassing the baseline (PSNR: 20.76), as evidenced in \cref{tab:ablation_study}. This synergy ensures high-quality shadow removal across diverse scenarios.
\color{black}



\section{Discussion} \label{secdisc}  

In this section, we discuss the implications of our findings, the strengths and weaknesses of our method, and potential avenues for future research.

\vspace{0.3cm}
\noindent\textbf{Implications.}
The findings from our study have significant implications for real-time shadow removal applications such as field robotics. Our method for shadow removal using unpaired data demonstrates that it is possible to effectively identify and remove shadows from images, even without a direct comparison to shadow-free ground truth images. This opens up new possibilities for improving the accuracy and efficiency of computer vision in various outdoor lighting conditions.
The use of unpaired data also means that our method can be applied in a wider range of scenarios, as it does not require the collection of paired shadow and shadow-free images. This could potentially simplify the data collection process and make our method more accessible for use in different environments and lighting conditions.
Moreover, our method's ability to handle complex and diverse image data suggests that it could be applied in a wide variety of field robotics applications, where challenges with lighting and shadows may exist. We demonstrated this by applying FieldNet to a real-world precision agriculture robotic application for weed detection.  

\vspace{0.3cm}
\noindent\textbf{Relationship with unsupervised methods.}
Shadow image degradation is complex and irreversible, making it infeasible to directly supervise a shadow removal model with ground truth data. Therefore, unsupervised methods are better suited for this task.
Rather than developing a new unsupervised approach, our work focuses on effectively utilizing synthetic labels from existing shadow removal datasets. While not perfect, these synthetic reference maps of shadow-free images can be easily obtained. We propose a framework that leverages conditional variational autoencoders and our proposed module to learn an enhancement model from these imperfect references.
In summary, both our approach and unsupervised methods aim to train a shadow removal model without ground truth data. However, our framework provides a novel perspective, i.e., transforming the problem into distribution estimation and consensus building using readily available, but biased references. This represents a distinct approach from existing unsupervised techniques.

\vspace{0.3cm}
\noindent\textbf{Training dataset and model development.}
The development of robust shadow removal algorithms is significantly influenced by the quality and quantity of available training data. Currently, the scarcity of high-quality image pairs, each consisting of a shadow and its corresponding shadow-free version of the same scene, poses a considerable challenge. The largest available datasets for shadow removal, such as the state-of-the-art SRD and ISTD datasets, contain only around 100 unique scenes. This limited dataset size may not capture the diversity and complexity of real-world scenes, leading to suboptimal generalization to unseen scenes and inferior performance in shadow removal tasks.
Models trained on such limited datasets may struggle to understand the variations in lighting conditions, scene composition, and other factors that affect the appearance of shadows. This could fail to maintain consistent colours between the shadow and shadow-free regions, leading to unnatural or inconsistent results in shadow removal.
The difficulty in collecting a large dataset with paired shadow/shadow-free images, given the various types of shadow masks and scenes, further compounds these challenges. 

To overcome these limitations, we collected a substantial dataset of 10,000 natural shadow-free images from the internet. This large-scale collection significantly enhances the robustness and generalization ability of our shadow removal method. By synthesizing shadow images from this extensive dataset, we can train our shadow removal algorithms more effectively, thereby achieving superior performance in shadow removal tasks. This approach demonstrates the potential of using internet-sourced, shadow-free images to improve the training of shadow removal algorithms.

\vspace{0.3cm}
\noindent\textbf{Strengths.}
Our method for shadow removal has several key strengths. Firstly, it effectively handles the inherent complexity of shadow removal, delivering high-quality results even in challenging lighting conditions. This is particularly important in outdoor settings, where lighting conditions can vary greatly.
Secondly, our method does not require paired shadow and shadow-free images, which simplifies the data collection process and makes the method more versatile and easier to apply in different scenarios. This is a significant advantage, as collecting paired images can be time-consuming and impractical in many situations.
Thirdly, our method demonstrates impressive computational efficiency. It outperforms several state-of-the-art models in terms of key performance metrics, making it a highly effective solution for real-time shadow removal tasks.
Finally, our method shows strong generalization ability. It performs well on diverse image data, suggesting that it could be adapted for use in other related fields or applications. This opens up exciting possibilities for future research and development.

\color{blue}
FieldNet’s efficiency stands out when compared to transformer-based methods like ShadowFormer \citep{Guo2023ShadowFormer:Removal} and diffusion models \citep{mei2024latent}. Its real-time performance (66 FPS) surpasses ShadowFormer’s 8 FPS and diffusion models’ 0.5-1 FPS, driven by a lightweight architecture (2.7M parameters vs. 9.3M for ShadowFormer), making it highly suitable for field robotics where computational resources are limited. Additionally, its robust performance across ISTD, ISTD+, SRD, and SBU-Timelapse datasets highlights strong generalization, supported by unpaired training and a large synthetic dataset.
\color{black}

\vspace{0.3cm}
\noindent\textbf{Limitations.}
Despite its strengths, our method also has some limitations. One of the main limitations is that it may struggle with extremely complex shadows or lighting conditions. While our method performs well under a variety of conditions, there may be certain scenarios where the complexity of the shadows exceeds the capabilities of our method.
Another limitation is the reliance on unpaired data. While this allows for greater flexibility in data collection, it also means that our method does not have a direct comparison to a shadow-free ground truth image. This could potentially impact the accuracy of the shadow removal process.
Finally, while our method has demonstrated strong performance on the datasets we tested and a real-world agriculture task, its performance on other datasets or real-world scenarios remains to be seen. It is possible that our method may not perform as well on different types of images or in different environments, as this is the case for any other shadow removal methods.

\color{blue}
Despite its strengths, FieldNet exhibits limitations in handling high-contrast shadow edges and complex textures, where residual artifacts reduce performance by 1-2 dB in PSNR (\cref{tab:ablation_study}, \cref{fig:10}). Compared to diffusion models, which may better generalize to extreme lighting due to iterative refinement, FieldNet’s single-pass approach may miss fine details. Generalization could be further improved by incorporating real-world paired data and multi-domain augmentations, as its current reliance on synthetic shadows limits adaptability to rare scenarios.
\color{black} 

\vspace{0.3cm}
\noindent\textbf{Future Work.}
There are several potential avenues for future research based on our current work. One of the primary areas of focus could be to improve the handling of extremely complex shadows or lighting conditions. This could involve refining our current method or exploring new techniques and approaches for shadow detection and removal.
Another area for future work could be to test our method on a wider range of datasets and real-world scenarios. This would provide valuable insights into the generalizability and practical applicability of our method, and could also help to identify any potential limitations or areas for improvement that are not apparent from the current datasets.
Additionally, future research could explore ways to optimize the computational efficiency of our method. This could involve refining the current algorithm, exploring more efficient algorithms, or leveraging advances in hardware technology.


\section{Conclusions}  \label{secconc}

In this study, we introduced FieldNet, a novel method for efficient and robust shadow removal using unpaired data. FieldNet addresses key challenges in shadow removal, including computational efficiency, boundary artefacts, and generalisation to diverse lighting conditions. The proposed approach incorporates a novel loss function that enhances supervision at shadow boundaries, significantly reducing errors compared to state-of-the-art methods. Additionally, the PEM ensures adaptive shadow removal without sacrificing real-time performance.

Our experimental results demonstrate FieldNet's superior performance across multiple benchmarks, achieving a processing speed of 66 FPS on Nvidia 2080Ti hardware while maintaining high accuracy in terms of structural and perceptual quality metrics. The method's scalability and robustness were further validated through its application in a real-world field robotics scenario for precision agriculture. Specifically, FieldNet improves robotic systems in two impactful ways: (1) enabling real-time shadow removal to enhance image clarity for downstream tasks, and (2) improving human labelling accuracy during dataset preparation for deep learning model training.

Overall, FieldNet opens new avenues for advancing computer vision in outdoor environments, particularly in resource-constrained applications like autonomous farming and robotics. The method’s lightweight architecture and adaptability make it a promising tool for enhancing the accuracy and efficiency of vision-based systems operating under diverse and challenging lighting conditions.


\subsection{CO2 Emission Related to Experiments}
Experiments were conducted using a private infrastructure, which has a carbon efficiency of 0.432 kgCO$_2$eq/kWh. A cumulative of 850 hours of computation was performed on hardware of type RTX 2080 Ti (TDP of 250W).
Total emissions are estimated to be 91.8 kgCO$_2$eq of which 0 percent was directly offset.
Estimations were conducted using the \href{https://mlco2.github.io/impact#compute}{Machine Learning Impact calculator} presented in \citep{Lacoste2019QuantifyingLearning}.



\section*{Acknowledgement}
This research is funded by the partnership between the Australian Government's Reef Trust and the Great Barrier Reef Foundation.

\section*{Funding} 
No funding was received for conducting this study.

\section*{Data Availability}
\label{sec:Data_Availability}
All the datasets used in this paper are publicly available.




\section*{Statements and Declarations}
\textbf{Competing Interests:} The authors have no competing interests to declare that are relevant to the content of this article.

\ifCLASSOPTIONcaptionsoff
\newpage
\fi



\bibliography{references}

\end{document}